\documentclass{article}

\usepackage[final,nonatbib]{neurips_2021}

\usepackage[utf8]{inputenc} 
\usepackage[T1]{fontenc}    
\usepackage[pagebackref=true,breaklinks=true,colorlinks,citecolor=gray]{hyperref}       
\usepackage{url}            
\usepackage{booktabs}       
\usepackage{amsfonts}       
\usepackage{nicefrac}       
\usepackage{microtype}      
\usepackage{xcolor}         

\title{Dynamic Visual Reasoning by \\ Learning Differentiable Physics Models \\from Video and Language}

\author{%
Mingyu Ding\\
MIT CSAIL and HKU\\
\And
Zhenfang Chen\\
MIT-IBM Watson AI Lab\\
\And
Tao Du\\
MIT CSAIL\\
\And
Ping Luo\\
HKU\\
\\
\And
Joshua B. Tenenbaum\\
MIT BCS, CBMM, CSAIL\\
\And
Chuang Gan\\
MIT-IBM Watson AI Lab\\
}

\usepackage{multirow}
\usepackage{booktabs}
\usepackage{subfigure}
\usepackage{amsmath,amsfonts,bm,amsthm,amssymb}
\usepackage{algorithm}
\usepackage{algorithmic}
\usepackage{graphics,graphicx,caption,float,color}
\usepackage{wrapfig}
\usepackage{bbm}
\usepackage{appendix}

\makeatletter
\newcommand*\bigcdot{\mathpalette\bigcdot@{.5}}
\newcommand*\bigcdot@[2]{\mathbin{\vcenter{\hbox{\scalebox{#2}{$\m@th#1\bullet$}}}}}
\makeatother

\definecolor{MyDarkBlue}{rgb}{0,0.08,1}
\definecolor{MyDarkGreen}{rgb}{0.02,0.6,0.02}
\definecolor{MyDarkRed}{rgb}{0.8,0.02,0.02}
\definecolor{MyDarkOrange}{rgb}{0.40,0.2,0.02}
\definecolor{MyPurple}{RGB}{111,0,255}
\definecolor{MyRed}{rgb}{1.0,0.0,0.0}
\definecolor{MyGold}{rgb}{0.75,0.6,0.12}
\definecolor{MyDarkgray}{rgb}{0.66, 0.66, 0.66}

\newcommand{\eg}{\emph{e.g.}}
\newcommand{\ie}{\emph{i.e.}}
\newcommand{\alias}{VRDP}
\newcommand{\full}{Visual Reasoning with Differentiable Physics}
\newcommand{\engine}{an impulse-based differentiable rigid-body simulator}

\begin{document}

\maketitle

\begin{abstract}
In this work, we propose a unified framework, called \full~(\alias)~\footnote{Project page: \url{http://vrdp.csail.mit.edu/}}, that can jointly learn visual concepts and infer physics models of objects and their interactions from videos and language.  
This is achieved by seamlessly integrating three components: a visual perception module, a concept learner, and a differentiable physics engine. The visual perception module parses each video frame into object-centric trajectories and represents them as latent scene representations. The concept learner grounds visual concepts (\eg, color, shape, and material) from these object-centric representations based on the language, thus providing prior knowledge for the physics engine.  The differentiable physics model, implemented as \engine, performs differentiable physical simulation based on the grounded concepts to infer physical properties, such as mass, restitution, and velocity, by fitting the simulated trajectories into the video observations. Consequently, these learned concepts and physical models can explain what we have seen and imagine what is about to happen in future and counterfactual scenarios.
Integrating differentiable physics into the dynamic reasoning framework offers several appealing benefits. More accurate dynamics prediction in learned physics models enables state-of-the-art performance on both synthetic and real-world benchmarks while still maintaining high transparency and interpretability; most notably, \alias~improves the accuracy of predictive and counterfactual questions by 4.5\% and 11.5\% compared to its best counterpart.  \alias~is also highly data-efficient: physical parameters can be optimized from very few videos, and even a single video can be sufficient. Finally, with all physical parameters inferred, \alias~can quickly learn new concepts from few examples. 

\end{abstract}

\section{Introduction}
Dynamic visual reasoning about objects, relations, and physics is essential for human intelligence. Given a raw video, humans can easily use their common sense of intuitive physics to explain what has happened, predict what will happen next, and infer what would happen in counterfactual situations.
Such human-like physical scene understanding capabilities are also of great importance in practical applications such as industrial robot control~\cite{ajay2019combining,li2018learning}.

Previous works have made great efforts to build artificial intelligence (AI) models with such physical reasoning capabilities. One popular strategy is to develop pure neural-network-based models~\cite{qi2020learning, ding2020object,hudson2018compositional}. These methods typically leverage end-to-end neural networks~\cite{he2016deep,HochSchm97} with powerful attention modules such as Transformer~\cite{vaswani2017attention} to extract attended features from both video frames and question words, based on which they answer questions directly. Despite their high question-answering accuracy on CLEVRER~\cite{yi2019clevrer}, a challenging dynamic visual question-answering benchmark, these black-box models neither learn concepts nor model objects' dynamics. Therefore, they lack transparency, interpretability, and generalizability to new concepts and scenarios.
Another common approach to dynamic visual reasoning is to build graph neural networks (GNNs)~\cite{kipf2017semi} to capture the dynamics of the scenes. These GNN models~\cite{li2019propagation,yi2019clevrer,chen2021grounding} treat objects in the video as nodes and perform object- and relation-centric updates to predict objects' dynamics in future or counterfactual scenes. Such systems achieve decent performance with good interpretability on CLEVRER by combining the GNN-based dynamics models with neural-symbolic execution~\cite{mao2019neuro,yi2018neural}. However, these dynamic models do not explicitly consider laws of physics or use concepts encoded in the question-answer pairs associated with the videos. As a result, they show limitations in counterfactual situations that require long-term dynamics prediction.   

Although (graph-)neural-network-based approaches have achieved competitive performance on CLEVRER, dynamic visual reasoning is still far from being solved perfectly. In particular, due to the lack of explicit physics models, existing models~\cite{yi2019clevrer,ding2020object,chen2021grounding} typically struggle to reason about future and counterfactual events, especially when training data is limited. For this reason, one appealing alternative is to develop explicit physics-based methods to model and reason about dynamics, as highlighted in the recent development of differentiable physics engines~\cite{battaglia2013simulation,de2018end,toussaint2018differentiable,degrave2019differentiable,smith2019modeling} and their applications in robotics~\cite{battaglia2013simulation,de2018end,toussaint2018differentiable}. However, these physics engines typically take as input a full description of the scene (\eg, the number of objects and their shapes) which usually requires certain human priors, limiting their availability to applications with well-defined inputs only.

In this work, we take an approach fundamentally different from either network-based methods or physics-based methods. Noting that deep learning based methods excel at parsing objects and learning concepts from videos and language, and physics laws are good at capturing object dynamics, we propose~\full~(\alias), a unified framework that combines a visual perception module, a concept learner, and a differentiable physics engine. \alias~jointly learns object trajectories, language concepts, and objects' physics models to make accurate dynamic predictions. It starts with a perception module running an object detector~\cite{he2017mask} on individual frames to generate object proposals and connect them into trajectories based on a motion heuristic. 
Then, a concept learner learns object- and event-based concepts, such as `shape', `moving', and `collision' as in DCL~\cite{chen2021grounding,mao2019neuro}.
Based on the obtained object trajectories and attributes, the differentiable physics engine estimates all dynamic and physical properties (\eg, velocity, angular velocity, restitution, mass, and the coefficient of resistance) by comparing the simulated trajectories with the video observations. With these explicit physical parameters, the physics engine reruns the simulation to reason about future motion and causal events, which a program executor then executes to get the answer. The three components of \alias~cooperate seamlessly: the concept learner grounds physical concepts needed by the physics engine like `shape' onto the objects detected by the perception module; the differentiable physics engine estimates all physical parameters and simulates accurate object trajectories, which in turn help the concept learning process in the concept learner.

Compared with existing methods, \alias~has several advantages thanks to its carefully modularized design. First, it achieves the state-of-the-art performance on both synthetic videos (CLEVRER~\cite{yi2019clevrer}) and real-world videos (Real-Billiard~\cite{qi2020learning}) without sacrificing transparency or interpretability, especially in situations that require long-term dynamics prediction. Second, it has high data efficiency thanks to the differentiable physics engine and symbolic representation.
Third, it shows strong generalization capabilities and can capture new concepts with only a few examples.

\section{Related Work}

\noindent \textbf{Visual Reasoning}~
Our model is related to reasoning on vision and natural language. 
Existing works can be generally categorized into two streams as end-to-end approaches~\cite{hudson2018compositional, zhu2016visual7w, Wu_2016_CVPR, hudson2019gqa,antol2015vqa} and neuro-symbolic approaches~\cite{yi2018neural, han2019visual, haninterpretable, gan2017vqs,andreas2016neural, mascharka2018transparency, johnson2017inferring, mao2019neuro, hu2018explainable, amizadeh2020neuro}.
The end-to-end methods~\cite{hudson2018compositional, zhu2016visual7w, Wu_2016_CVPR, misra2018learning} typically tackle the visual question answering (VQA) problem by designing monolithic multi-modal deep networks~\cite{he2016deep,HochSchm97}. They directly output answers without explicit and interpretable mechanisms.
Beyond question answering, neuro-symbolic methods~\cite{yi2018neural, han2019visual, mascharka2018transparency, johnson2017inferring, mao2019neuro} propose a set of visual-reasoning primitives, which manifest as an attention mechanism capable of performing complex reasoning tasks in an explicitly interpretable manner.

Dynamic visual reasoning in videos has attracted much research attention.
Many video question answering datasets~\cite{girdhar2019cater,lei2018tvqa,jang2017tgif,MovieQA,zadeh2019social} and the methods~\cite{zhou2021hopper,li2019beyond,huang2020location,xu2017video,ye2017video,fan2019heterogeneous} built on them mainly focus on understanding diverse visual scenes, such as human actions (MovieQA~\cite{MovieQA}) or 3D object movements without physical and language cues (CATER~\cite{girdhar2019cater}).
Differently, CLEVRER~\cite{yi2019clevrer} targets the physical and causal relations grounded in dynamic videos of rigid-body collisions and asks a range of questions that requires the modeling of long-term dynamic predictions. For this reason, we evaluate our method and compare it with other state-of-the-arts on CLEVRER.

Both end-to-end~\cite{qi2020learning, ding2020object} and neuro-symbolic methods~\cite{yi2019clevrer,chen2021grounding} have been explored on CLEVRER.
However, they either lack transparency or struggle for long-term dynamic prediction.
In this paper, we perform high-performance and interpretable reasoning by recovering the physics model of objects and their interactions (\eg, collisions) from visual perception and language concepts. 

\noindent \textbf{Physical Models}~
Physical models are widely used in video prediction~\cite{lerer2016learning,finn2016unsupervised,ye2018interpretable,ye2019compositional}, neural simulation and rendering~\cite{li2020visual,li2019propagation}, and dynamic reasoning~\cite{battaglia2016interaction,watters2017visual}. For example,  PhysNet and its variants~\cite{lerer2016learning,finn2016unsupervised,ye2018interpretable,ye2019compositional} leverage global or object-centric deep features to predict the physical motion in video frames. Some other related works~\cite{mottaghi2016happens,agrawal2016learning,shao2014imagining,kipf2019contrastive} extend physical models to predict the effect of forces and infer the geometric attributes and topological relations of cuboids.

In this work, we focus on dynamic visual reasoning about object interactions, dynamics, and physics with question answering, which is central to human intelligence and a key goal of artificial intelligence. Solving such tasks requires a good representation and understanding of physics models. A common choice is to train a deep neural network for physical property estimation (\eg, location and velocity) based on learned visual and dynamic priors~\cite{chang2016compositional, battaglia2016interaction, watters2017visual, veerapaneni2020entity, janner2018reasoning, li2019propagation, liang2020differentiable,heiden2019interactive,heiden2020neuralsim}. However, since these neural networks do not model physics laws, generalizing them to unseen events or counterfactual scenarios could result in unexpected results. Our work is different and more physics-based: Inspired by the recent advances in differentiable physics~\cite{battaglia2013simulation,de2018end,toussaint2018differentiable,degrave2019differentiable,smith2019modeling,hu2019difftaichi}, we implement \engine~and leverage the power of its gradients to infer dynamics information about the scene.

\noindent \textbf{Physical Scene Understanding}~
Our work is also relevant to studies on physical scene understanding~\cite{bakhtin2019phyre,galileo,smith2019modeling,riochet2018intphys,baradel2019cophy,gan2020threedworld,gan2021transport,lerer2016learning,huang2021plasticinelab}, most of which propose pure neural-network solutions without explicitly incorporating physics models. Benchmarks like PHYRE~\cite{bakhtin2019phyre} study physical understanding and reasoning based on pure videos without concept learning and language inference.
Based on such benchmarks, some works~\cite{wu2017learning, bear2020learning, lerer2016learning, li2019propagation} learn compositional scene structure and estimate states through physical motions and visual de-animation. Recently, two papers propose pure physics-based methods~\cite{jatavallabhula2021gradsim,huang2021plasticinelab} that make heavy use of differentiable physics simulators, but they typically assume concepts in the scene are given as input. Our work is unique in that we learn video and language concepts from raw videos and infer dynamics information from a differentiable simulator, combining the benefits of both learning and physics.

\begin{figure}[t]
    \centering
    \includegraphics[width=1\linewidth]{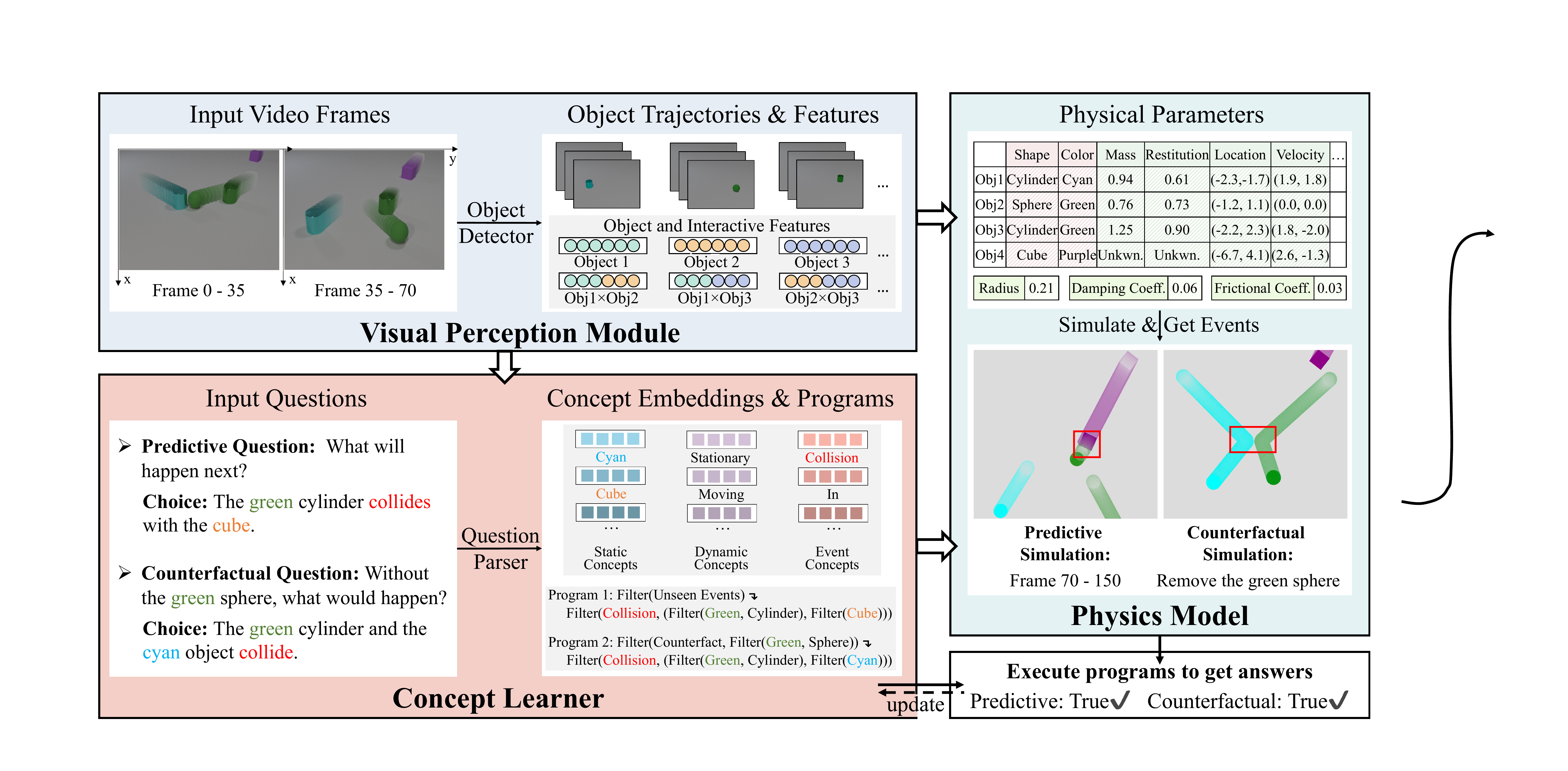}
    \caption{\alias~contains three components including a visual perception module, a concept learner, and a physics model.
    The perception module first runs an object detector~\cite{he2017mask} on individual frames to generate object proposals and connect them into trajectories based on motion heuristic. 
    Then, the concept learner learns object- and event-based concepts, such as `shape', `moving', and `collision', as prior knowledge for the physics model.
    Based on the obtained object trajectories and concepts, the differentiable physics engine estimates all dynamic and physical properties (\eg, velocity $v$, angular velocity $\alpha$, restitution $r$, mass $m$, and coefficients of resistance $\lambda$) by comparing the simulated trajectories with the video observations.
    With these explicit physical parameters, the physic engine reruns the simulation to reason about future motion and causal events, which are then executed by a symbolic executor to get the answer.
    Stroboscopic imaging is applied for motion visualization.
}
    \label{fig:overview}
    \vspace{-5pt}
\end{figure}

\section{Method}

By integrating differentiable physics into the dynamic reasoning framework, \alias~jointly learns visual perception, language concepts, and physical properties of objects. The first two provide prior knowledge for optimizing the third one to reason about the physical world, and the optimized physical properties in turn help to learn better concepts. In the following, we first give an overview of our framework and then describe each of its components in detail.

\vspace{-1pt}
\subsection{Framework Overview}
An overview of \alias~is illustrated in Fig.~\ref{fig:overview}. It contains three components: a visual perception module, a concept learner, and a physics model. 
The input to the framework is a video and reasoning questions, where the former is processed by the visual perception module to get object trajectories and corresponding visual features, and the latter is parsed into executable symbolic programs with language concepts by the concept learner.
Similar to DCL~\cite{mao2019neuro,chen2021grounding}, the concept learner first grounds object properties (\eg, color and shape) and event concepts (\eg, collision and moving) by aligning the visual features and the corresponding concept embeddings in the (explanatory or descriptive) program that does not require dynamic predictions, \eg, ``what is the shape of ...''.
With those perceptually grounded object trajectories and properties, the physical model then performs differentiable simulation to learn all physical parameters of the scene and objects by comparing the simulated trajectories with the video observations.
After that, the physics engine simulates unseen trajectories for predictive and counterfactual scenarios and generates their features, in turn enabling the concept learner to finetune event concepts from the program that requires dynamic predictions, \eg, ``what will happen ...'' and ``what if ...''.
Finally, a symbolic executor executes the parsed programs with the dynamic predictions to get the answer.

\subsection{Model Details}
\noindent \textbf{Visual Perception Module}~
Given a video with the number of frames $T$, the visual perception module parses the video frame-by-frame and associate the parsed objects in each frame into object trajectories $L=\{l^n\}^N_{n=1}$, where $l^n$ denotes the object trajectory of the $n^{th}$ object and $N$ is the number of the objects in the video.
Specifically, we leverage a pretrained Faster R-CNN~\cite{he2017mask} as the object detector to get the Region of Interest (ROI) feature $f_t \in \mathbb{R}^{N\times D}$ and the object location of objects $b_t=[x^\text{2D}_t, y^\text{2D}_t, x^\text{BEV}_t, y^\text{BEV}_t] \in \mathbb{R}^{N\times 4}$ at frame $t$, where $D$ is the feature dimension, $(x^\text{2D}_t, y^\text{2D}_t)$ denotes the normalized object bounding box center in the image coordinate frame, and $(x^\text{BEV}_t, y^\text{BEV}_t)$ denotes the projected bird's-eye view (BEV) location in the BEV coordinate frame using the calibrated camera matrix.
Following works~\cite{chen2021grounding,gkioxari2015finding}, we associate object proposals in adjacent frames by thresholding their intersection over union (IoU) and obtain the object trajectory $l^n=\{b^n_t\}^T_{t=1}$ for the $n^{th}$ object.

The visual perception module then constructs object and interactive representations for concept learning. The object representation $F_\text{obj} \in \mathbb{R}^{N\times (D+4T)}$ contains both appearance-based $F_a=\mathrm{avg}(\{f_t\}^T_{t=1})$ and trajectory-based feature $F_l=\{b_t\}^T_{t=1}$ for modeling static properties and dynamic concepts, respectively, where $\mathrm{avg}(\cdot)$ here represents the average ROI feature over time. 
The interactive feature $F_\text{pair} \in \mathbb{R}^{T\times N\times N\times 12S}$, where $S$ denotes a fixed temporal window size, is built on every pair of objects.
It contains object trajectories $\{b_t^i\}_{t_0-S/2}^{t_0+S/2}, \{b_t^j\}_{t_0-S/2}^{t_0+S/2}$ of the objects $i$ and $j$ and their distance $\{\mathrm{abs}(b_t^i - b_t^j)\}_{t_0-S/2}^{t_0+S/2}$ to model the collision event of the objects at a specific moment $t_0$.

\noindent \textbf{Concept Learner}~
The concept learner grounds the physical and event concepts (\eg, shape and color) as prior knowledge for the physics model from the video representation and language.
It first leverages a question parser to translate the input questions and choices into executable neuro-symbolic programs where each language concept in the program is represented by a concept embedding.
Similar to \cite{mao2019neuro,chen2021grounding}, this work adopts a seq2seq model~\cite{bahdanau2014neural} with an attention mechanism to translate language into a set of symbolic programs, \eg, retrieving objects with certain colors, getting future or counterfactual events, finding the causes of an event, thus decomposing complex questions into step-by-step dynamic visual reasoning processes.
The concept learner assigns each concept in the program (\eg, color, shape, and collision) a randomly initialized embedding $e \in \mathbb{R}^C$ so that the symbolic program can be formulated as differentiable vector operations.

After that, it projects the visual representation into concept embedding spaces and performs Nearest Neighbor (NN) search to quantize concepts for the object attributes and events.
We implement the projection through a linear layer $\mathcal{P}(\cdot)$ and calculate the cosine similarity between two vectors in the embedding space for NN search. For example, the confidence score of whether the $n^{th}$ object is a cube can be represented by $[\mathrm{cos}(\mathcal{P}(F_a^{n}), e_\text{cube}) -\mu]/\sigma$, where $e_\text{cube}$ is the embedding of concept `cube', $\mu$ and $\sigma$ are the shifting and scaling scalars, and $\mathcal{P}(\cdot)$ maps a $D$-dimensional visual feature into a $C$-dimensional vector in this case.


\noindent \textbf{Physics Model}~
The differentiable physics model captures objects' intrinsic physical properties and makes accurate dynamic predictions for reasoning. With the perceptually grounded object shapes and trajectories from the above two components of \alias, it performs differentiable simulation to optimize the physical parameters of the scene and objects by comparing the simulation with the video observations $L$.
Our physics model is implemented as \engine~\cite{hu2019difftaichi,muller2008real,catto2009modeling}. It iteratively simulates a small time step of $\Delta t$ based on the objects' state in the BEV coordinate through inferring collision events, forces (including resistance and collision force), and impulses acting on the object, and updating the state of each object.

When an object moves on the ground with velocity $\overrightarrow{v}$ and angular velocity $\omega$, we consider three kinds of forces that affect the movement of the object: sliding friction, rolling resistance, and air resistance. We use $\lambda_1, \lambda_2, \lambda_3$ to denote their coefficients and have:
\begin{equation}
\overrightarrow{a} = 
\begin{cases}
-\frac{\overrightarrow{v}}{|\overrightarrow{v}|}(\lambda_1 \mathrm{g} + \lambda_3 |\overrightarrow{v}|^2)~~~~~\text{if the shape is not sphere} \\
-\frac{\overrightarrow{v}}{|\overrightarrow{v}|}(\lambda_2 \mathrm{g} + \lambda_3 |\overrightarrow{v}|^2)~~~~~\text{if the shape is sphere}
\end{cases}
\label{eq:resistance}
\end{equation}
where $\mathrm{g}=9.81\mathrm{m/s^2}$ is the standard gravity and $\overrightarrow{a}$ denotes the acceleration of the object, whose direction is opposite to the velocity. The velocity $\overrightarrow{v}$ and the location $\overrightarrow{l'}=(x', y')$ are then updated accordingly by the second order Runge-Kutta (RK2) algorithm~\cite{butcher1975stability}.
Similarly, the angular velocity $\omega$ also decreases at each time step due to the angular drag, and the angle $\alpha$ of the object is updated by the RK2 algorithm.

The physics engine checks whether the boundaries of two objects with radius $R$ are overlapped in the BEV coordinate frame to detect collision events.
Based on the fact that the total momentum of an isolated system should be constant in the absence of net external forces, we compute the impulse of collided objects and ignore the friction caused by the collision.
Let $(m_1, m_2)$, $(r_1, r_2)$, $(\alpha_1$, $\alpha_2)$, $(\overrightarrow{v_1}, \overrightarrow{v_2})$, $(\overrightarrow{l'_1}, \overrightarrow{l'_2})$ denote the mass, restitution, angle, velocity and BEV location of two collided objects at the moment of the collision, respectively; $\overrightarrow{d_1}, \overrightarrow{d_2}$ represent their collision unit directions that the force is acting on, where $\overrightarrow{d_1} + \overrightarrow{d_2} = \overrightarrow{0}$. The change of velocity $\Delta\overrightarrow{v_1}, \Delta\overrightarrow{v_2}$ at the moment of collision can be obtained by calculating the impulse on the collision direction:
\begin{equation}
\begin{split}
& \overrightarrow{\Delta v_1} = -(1 + r_1 r_2) (m_2 / (m_1 + m_2)) (\overrightarrow{d_1} \bigcdot (\overrightarrow{v_1} - \overrightarrow{v_2})) \overrightarrow{d_1} \\
& \overrightarrow{\Delta v_2} = -(1 + r_1 r_2) (m_1 / (m_1 + m_2)) (\overrightarrow{d_2} \bigcdot (\overrightarrow{v_2} - \overrightarrow{v_1})) \overrightarrow{d_2},
\end{split}
\label{eq:dynamics}
\end{equation}
the velocity $\overrightarrow{v}$ is then updated by $\overrightarrow{v} \leftarrow \overrightarrow{v} + \Delta \overrightarrow{v}$.
Similarly, the angular velocity $\omega$ can be updated by $\omega \leftarrow \omega + \Delta \omega$, where $\Delta \omega$ is computed based on conservation of angular momentum.

Given an initial state of the scene and objects, our physics engine simulates force, impulse, and collision events and iteratively updates the state of each element. 
All physical parameters including $R, \lambda, m, r, \alpha, \overrightarrow{v}, \overrightarrow{l}$ are initialized and then optimized with L-BFGS algorithm~\cite{liu1989limited} by fitting the simulated trajectories $L'=\{(x'_t, y'_t)\}_{t=1}^T$ into the perceptual trajectories $L^\text{BEV} = \{(x^\text{BEV}_t, y^\text{BEV}_t)\}_{t=1}^T$.
To alleviate the difficulty of the optimization, we mark the time frame of each object's first collision by calculating the BEV distance between every pair and decompose the differentiable physical optimization and simulation into the following steps:
1) Since radius $R$ and resistance coefficients $\lambda$ are consistent in all videos, we use $K$ videos to jointly learn those physical parameters and fix them for the optimization of other sample-dependent parameters.
2) For each video, we then use the frames before the collision to optimize the collision-independent physical parameters, such as initial velocity $\overrightarrow{v_0}$, initial location $\overrightarrow{l_0}$, and initial angle $\alpha_0$.
3) With the above parameters learned and fixed, we optimize the remaining collision-dependent parameters, including mass $m$ and resistance $r$ of each object. This process follows the curriculum learning paradigm~\cite{bengio2009curriculum} by optimizing from fewer to more frames, \eg, multi-step optimization on [0, 40], [0, 80], and [0, 128] frames, where the parameters in each step are initialized from the optimization of the previous step.
4) With all parameters of the physical model learned, the engine runs simulations and re-calculates the trajectory-based representations $F_l$ for answering counterfactual, descriptive, and explanatory questions.
5) For the predictive case, we leverage the learned physical model as initialization and re-optimize all sample-dependent parameters with only the last 20 frames to reduce the cumulative error over time.

\noindent \textbf{Symbolic Execution}~
As in \cite{mao2019neuro,chen2021grounding}, we perform reasoning with a program executor, which is a collection of deterministic functional modules designed to realize all logic operations specified in symbolic programs. Its input consists of the parsed programs, learned concept embeddings, and visual representations, including the appearance-based feature $F_a$ from the visual perception module and the updated trajectory feature $F_l$ from the physics engine.
Given a set of parsed programs, the program executor runs them step-by-step and derives the answer based on these representations. For example, the `counting' program outputs the number of objects which meet specific conditions (\eg, red sphere). In this process, the executor leverages the concept learner to filter out eligible objects.

Our reasoning process is designed fully differentiable w.r.t. the visual representations and the concept embeddings by representing all object states, events, and results of all operators in a probabilistic manner during training, supporting gradient-based optimization. Moreover, it works seamlessly with our explicit physics engine, which simulates dynamic predictions through real physical parameters, forming a symbolic and deterministic physical reasoning process. The whole reasoning process is fully transparent and step-by-step interpretable.

\subsection{Training Objectives}

Similar to \cite{chen2021grounding,yi2019clevrer}, we train the program parser with program labels using cross-entropy loss,
\begin{equation}
    \mathcal{L}_{program} = -\sum_{j=1}^J  \mathbf{1}\{y_p=j\}\log(p_j), 
\end{equation}
where $J$ is the size of the pre-defined program set, $p_j$ is the probability for the $j$-th program and $y_p$ is the ground-truth program label.

We optimize the physical parameters in the physical model by comparing the simulation trajectories with the video observations. All physical parameters including $R, \lambda, m, r, \alpha, \overrightarrow{v}, \overrightarrow{l}$ are initialized and then optimized with L-BFGS algorithm~\cite{liu1989limited} by fitting the simulated trajectories $L'=\{(x'_t, y'_t)\}_{t=1}^T$ into the perceptual trajectories $L^\text{BEV} = \{(x^\text{BEV}_t, y^\text{BEV}_t)\}_{t=1}^T$. We have:
\begin{equation}
    \mathcal{L}_{Physics} = \|L' - L^\text{BEV}\|^2_2,
\end{equation}

We optimize the feature extractor and the concept embeddings in the concept learner by question answering. We treat each option of a multiple-choice question as an independent boolean question during training.
Specifically, we use cross-entropy loss to supervise open-ended questions and use mean square error loss to supervise counting questions. Formally, for counting questions, we have
\begin{equation}
    \mathcal{L}_{QA, count} = (y_a - z)^2, 
\end{equation}
where $z$ is the predicted number and $y_a$ is the ground-truth number label.
For other open-ended questions and multiple-choice questions, we have
\begin{equation}
    \mathcal{L}_{QA, others} = -\sum_{a=1}^A  \mathbf{1}\{y_a=a\}\log(p_a),
\end{equation}
where $A$ is the size of the pre-defined answer set, $p_a$ is the probability for the $a$-th answer and $y_a$ is the ground-truth answer label. 

\section{Experiments}
By recovering physics models of objects and their interactions from video and language, \alias~enjoys the following benefits: 1) high accuracy and full transparency, 2) superior data efficiency, and 3) high generalizability. 
In this section, we first evaluate the accuracy and data efficiency of \alias~on the widely used dynamic visual reasoning benchmark CLEVRER~\cite{yi2019clevrer} and its subsets; we then validate the model's generalizability on adapting to new concepts with few-shot data; lastly, we experiment on the real-world dataset Real-Billiard~\cite{qi2020learning} to show that \alias~works well in real-world dynamic prediction and reasoning. 

\noindent \textbf{Datasets and Evaluation Settings}~~
To validate the effectiveness of our method for reasoning about the physical world, we conduct main experiments on the CLEVRER~\cite{yi2019clevrer} dataset, as it contains both language and physics cues such as rigid body collisions and dynamics, compared to other benchmarks that focus on either action understanding without physical inferring~\cite{lei2018tvqa,jang2017tgif} or temporal reasoning without language concepts~\cite{girdhar2019cater, baradel2019cophy}.
CLEVRER includes four types of question answering (QA): descriptive, explanatory, predictive, and counterfactual, where the first two types concern more on video understanding, while the latter two types involve physical dynamics and predictions in reasoning. Therefore, we mainly focus on the predictive and counterfactual questions in this work and use QA accuracy as the evaluation metric.
Note the multi-choice question (explanatory, predictive, and counterfactual) contains multiple options. Only if all the options (per opt.) are answered correctly can it be regarded as a correct question (per ques.).

We then collect a few-shot physical reasoning dataset with novel language and physical concepts (\eg, ``heavier'' and ``lighter''), termed generalized CLEVRER, containing 100 videos (split into 25/25/50 for train/validation/test) with 375 options in 158 counterfactual questions. This dataset is supplementary to CLEVRER~\cite{yi2019clevrer} for generalizing to new concepts with very few samples. For real-world scenarios, we conduct experiments on the Real-Billiard~\cite{qi2020learning} dataset, which contains three-cushion billiards videos captured in real games for dynamics prediction. We generate 6 reasoning questions (\eg, ``will one billiard collide with ...?'') for each video and evaluate both the prediction error and QA accuracy.

\noindent \textbf{Implementation Details}~~
We follow the experimental setting in \cite{yi2019clevrer,chen2021grounding} using a pre-trained Faster R-CNN model~\cite{he2017mask} to generate object proposals for each frame and training the language program parser with 1,000 programs for all question types.
We implement three versions of VRDP models, where our unsupervised \alias~leverage a Slot-Attention model~\cite{locatello2020object} to parse the objects unsupervisedly,
while \alias~$\dag$ use Faster-RCNN~\cite{he2017mask} as the object detector.
In addition to our standard model that grounds object properties from question-answer pairs, we also train a variant (\alias~$\dag\ddag$) on CLEVRER with an explicit rule-based program executor~\cite{yi2019clevrer} and object attribute supervision. The camera matrix is optimized from 20 training videos. We set $\Delta t=0.004\text{s}, K=10, S=10$, and $T=128$ for CLEVRER~\cite{yi2019clevrer} and $T=20$ for Real-Billiard~\cite{qi2020learning}. 
More details of the dataset and settings can be found in Supplemental Materials.

\begin{table}[t]
	\begin{center}
	\small
	\setlength{\tabcolsep}{5pt}
	\resizebox{1\linewidth}{!}{%
	\begin{tabular}{lccccccccc}
	\toprule
        \multirow{2}{*}{Methods} & \multicolumn{2}{c}{Overall} & \multicolumn{2}{c}{Predictive} & \multicolumn{2}{c}{Counterfactual} & \multirow{2}{*}{Descriptive} & \multicolumn{2}{c}{Explanatory} \\ 
    \cmidrule(lr){2-3}\cmidrule(lr){4-5}\cmidrule(lr){6-7}\cmidrule(lr){9-10}
        & per task & per ques. & per opt. & per ques. & per opt. & per ques. &  & per opt. & per ques. \\ 
    \midrule
        \midrule
        TVQA+~\cite{lei2018tvqa} & 37.2 & 57.3 & 70.3 & 48.9 & 53.9 & 4.1 & 72.0 & 63.3 & 23.7\\    
        Memory~\cite{fan2019heterogeneous} & 27.2 & 43.3 & 50.0 & 33.1 & 54.2& 7.0 & 54.7 & 53.7 & 13.9 \\
        IEP (V)~\cite{johnson2017inferring} & 20.2 & 40.5 & 50.0 & 9.7 & 53.4 & 3.8 & 52.8 & 52.6 & 14.5  \\ 
        TbD-net (V)~\cite{mascharka2018transparency} &23.6 & 58.6 & 50.3 & 6.5 & 56.1 & 4.4 & 79.5 & 61.6 & 3.8 \\
        HCRN~\cite{le2020hierarchical} & 27.3 & 44.8 & 54.1 & 21.0 & 57.1 & 11.5 & 55.7 & 63.3 & 21.0 \\ 
        MAC (V)~\cite{hudson2018compositional} & 32.1 & 65.5 & 51.0 & 16.5 & 54.6 & 13.7 & 85.6 & 59.5 & 12.5 \\ 
        MAC (V+)~\cite{hudson2018compositional} $^{\dag}$ & 44.2 & 69.8 & 59.7 & 42.9 & 63.5 & 25.1 & 86.4 & 70.5 & 22.3 \\ 
        NS-DR~\cite{yi2019clevrer} $^{\dag\ddag}$ & 69.7 & 80.7 & 82.9 & 68.7 & 74.1 & 42.2 & 88.1 & 87.6 & 79.6 \\ 
        NS-DR (NE)~\cite{yi2019clevrer} $^{\dag\ddag}$ & 64.1 & 77.7 & 75.4 & 54.1 & 76.1 & 42.0 & 85.8 & 85.9 & 74.3 \\
        DCL~\cite{chen2021grounding} $^{\dag}$ & 75.5 & 84.1 & 90.5 & 82.0 & 80.4 & 46.5 & 90.7 & 89.6 & 82.8 \\
        DCL-Oracle~\cite{chen2021grounding} $^{\dag\ddag}$ & 75.6 & 84.5 & 90.6 & 82.1 & 80.7 & 46.9 & 91.4 & 89.8 & 82.0 \\
        Object-based Attention~\cite{ding2020object} & 88.3 & 91.7 & 93.5 & 87.5 & 91.4 & 75.6 & \textbf{94.0} & \textbf{98.5} & \textbf{96.0} \\
        \midrule
        VRDP (ours) & 82.9 & 86.9 & 91.7 & 83.8 & 89.9 & \textbf{75.7} & 89.8 & 89.1 & 82.4 \\
        VRDP (ours) $^{\dag}$ & 86.6 & 89.4 & \textbf{94.5} & \textbf{89.2} & \textbf{92.5} & \textbf{80.7} & 91.5 & 90.9 & 85.2 \\
        VRDP (ours) $^{\dag\ddag}$ & \textbf{90.3} & \textbf{92.0} & \textbf{95.7} & \textbf{91.4} & \textbf{94.8} & \textbf{84.3} & 93.4 & 96.3 & 91.9 \\
    \bottomrule
	\end{tabular}}
	\end{center}
	\caption{Question-answering accuracy of visual reasoning models on CLEVRER~\cite{yi2019clevrer}. We report per-task and per-question overall accuracies, as well as per-option and per-question accuracies for each sub-task. Note that predictive and counterfactual questions that require dynamics and physical prediction are our focus.
	$^\dag$ denotes the method uses a supervised object detector, such as Faster/Mask R-CNN~\cite{he2017mask}.
	$^\ddag$ indicates the use of object properties (\ie, shape, color, and material) as supervision.}
	\label{tab:comparsion_clevrer}
	\vspace{-10pt}
\end{table}

\subsection{Comparative Results on CLEVRER}
We conduct experiments on CLEVRER against several counterparts: TVQA+~\cite{lei2018tvqa}, Memory~\cite{fan2019heterogeneous}, IEP (V)~\cite{johnson2017inferring}, TbD-net (V)~\cite{mascharka2018transparency}, HCRN~\cite{le2020hierarchical}, MAC~\cite{hudson2018compositional}, NS-DR~\cite{yi2019clevrer}, DCL~\cite{chen2021grounding}, and Object-based Attention~\cite{ding2020object}. Among them, NS-DR~\cite{yi2019clevrer} and DCL~\cite{chen2021grounding} are high-performance interpretable symbolic models, while Object-based Attention~\cite{ding2020object} is the state-of-the-art end-to-end method.

From Tab.~\ref{tab:comparsion_clevrer} we observe that:
1) Counterfactual and predictive questions are more difficult than descriptive and explanatory ones as they require accurate physical dynamics and prediction hence our main focus. By reconstructing the physical world explicitly, our method outperforms all existing works on these two types by large margins.
For example, \alias~$\dag$ improves the per question accuracy of counterfactual questions by 11.5\% and 79.7\% compared to the best end-to-end~\cite{ding2020object} and neural-symbolic~\cite{chen2021grounding} counterparts.

2) The end-to-end model~\cite{ding2020object} improves the accuracy at the cost of losing model transparency and interpretability. 
However, by leveraging object attribute supervision and explicit program executors~\cite{yi2019clevrer}, our \alias~$\dag\ddag$ achieves new state-of-the-art overall performance on CLEVRER. It closes the performance gap between interpretable models and state-of-the-art end-to-end methods. Moreover, it shows the flexibility of our physics model that can be combined with various physical concepts and program executors while achieving impressive performance.

3) We conducted ablative experiments to study the impact of pre-trained object detection modules of our framework by replacing the supervised visual model~\cite{he2017mask} in \alias~$\dag$ with an unsupervised one~\cite{locatello2020object} in \alias. We observe that although the use of unsupervised detectors decreases the performance slightly, our framework still enjoys higher performance than previous methods in counterfactual and predictive questions.

4) Neither the neuro-symbolic nor end-to-end works employ explicit dynamic models with physical meanings. In contrast, our model is fully transparent with step-by-step interpretable programs and meaningful physical parameters powered by a differentiable engine.

\begin{figure}[t]
    \centering
    \includegraphics[width=\linewidth]{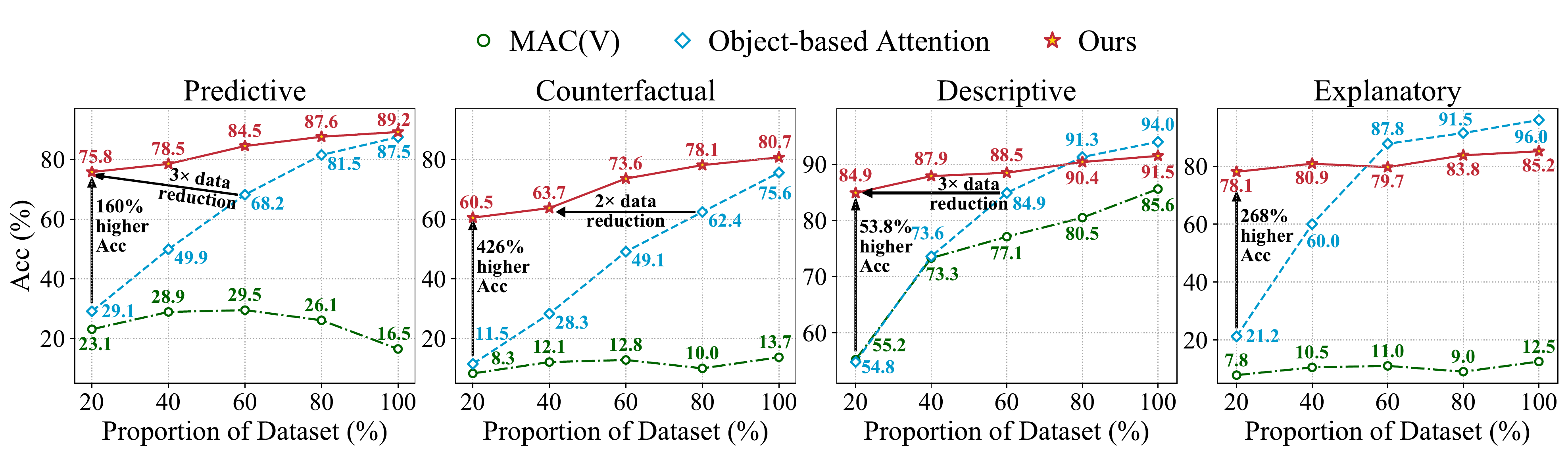}
    \caption{Comparisons of the data efficiency evaluation on four types of questions with MAC (V)~\cite{hudson2018compositional} and Object-based Attention~\cite{ding2020object} trained with different proportion of the CLEVRER~\cite{yi2019clevrer} dataset. Our method is highly data-efficient in that it achieves comparable results with the state-of-the-art counterpart~\cite{ding2020object} with $3\times$ fewer data. It improves the reasoning accuracy significantly when fewer data (\eg, 20\%) are used.}
    \label{fig:data_efficiency}
    \vspace{-5pt}
\end{figure}

\subsection{Detailed Analysis}

\noindent \textbf{Evaluation of Data Efficiency}~~
We evaluated the data efficiency of \alias~with two representative models: MAC (V)~\cite{hudson2018compositional} and Object-based Attention~\cite{ding2020object}. From Fig.~\ref{fig:data_efficiency} we see that:
\alias~is highly data-efficient. When the amount of data is reduced, the accuracy of our model drops slightly, while the performance of MAC (V)~\cite{hudson2018compositional} and Object-based Attention~\cite{ding2020object} drops drastically due to insufficient data.
For example, we improve the counterfactual accuracy of Object-based Attention~\cite{ding2020object} by 426\% under the setting of 20\% data. Notably, our model uses 20\% of the dataset to achieve comparable performance to other works that use 80\% of data.
This is because the components of \alias, \eg, perception module and question parser, can be trained with a small amount of data. More importantly, our physics model is built based on an explicit physics engine, which can be optimized from the trajectory of a single video.

\begin{figure}[t]
\begin{minipage}[t]{0.46\linewidth}
\begin{figure}[H]
    \centering
    \includegraphics[width=\linewidth]{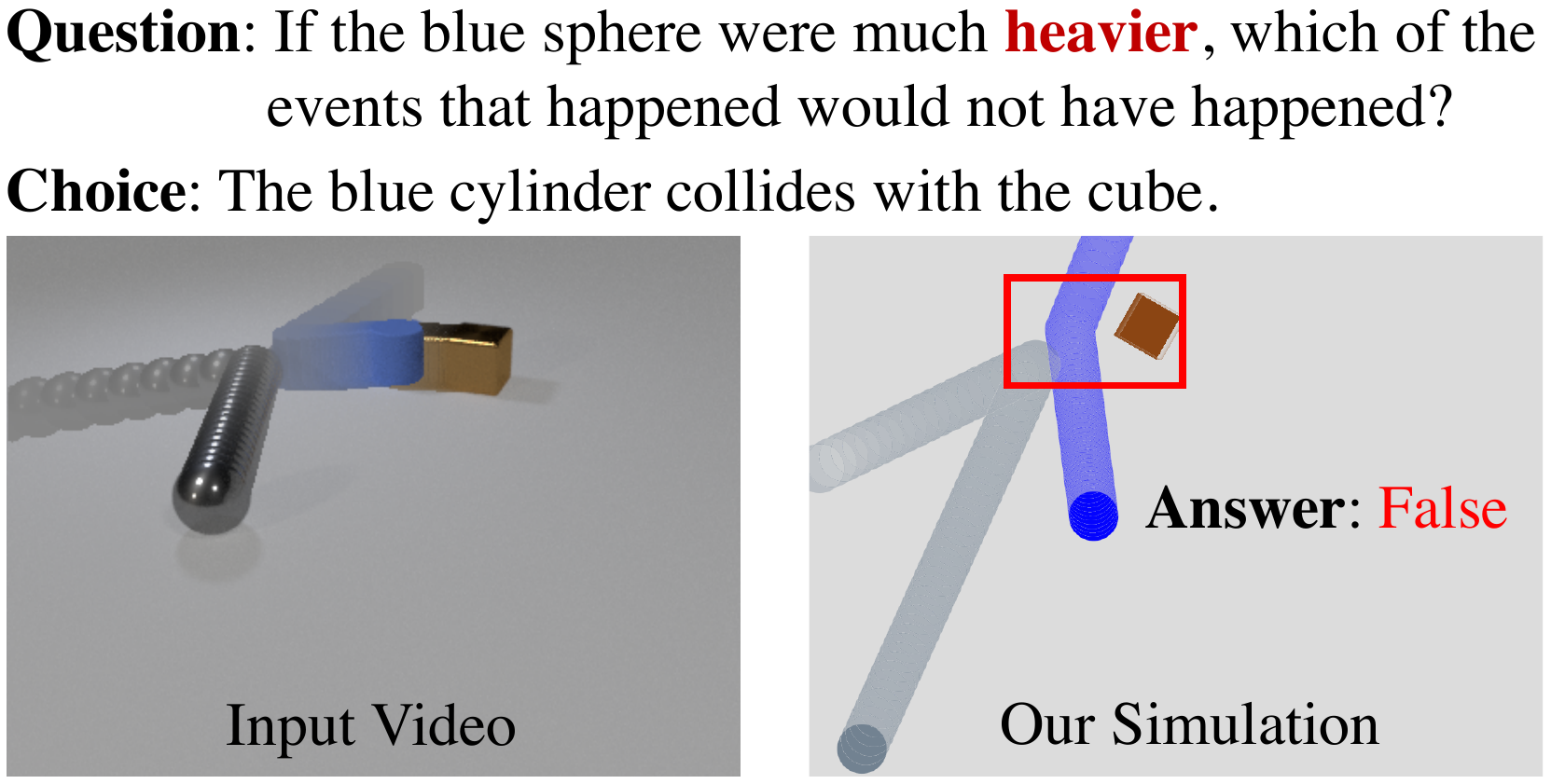}
    \vspace{-16pt}
    \caption{\alias~learns new concepts and accurately reasons about counterfactual events from few data on generalized CLEVRER.}
    \label{fig:few-shot}
\end{figure}
\end{minipage}
\hfill
\begin{minipage}[t]{0.51\linewidth}
\begin{table}[H]
\centering
\setlength{\tabcolsep}{2pt}
\resizebox{1\linewidth}{!}{%
    \begin{tabular}{lcc}
    \toprule
    Methods & Per opt. & Per ques. \\
    \midrule
    MAC (V)~\cite{hudson2018compositional} & 63.8 & 22.0 \\
    Object-based Attention~\cite{ding2020object} & 59.5 & 26.7 \\
    \alias~(Ours) & \textbf{88.1} & \textbf{75.6}  \\
    \bottomrule
    \end{tabular}
}
\vspace{2pt}
\caption{Comparative results of generalizability evaluation under the few-shot setting. All models are first pretrained on CLEVRER and then finetuned with only 25 videos for adapting to generalized CLEVRER. \alias~can learn new concepts quickly with few-shot data.}
\label{tab:few-shot}
\end{table}
\end{minipage}
\vspace{-8pt}
\end{figure}

\noindent \textbf{Evaluation of Generalizability}~~
This part studies the generalization capabilities of \alias~against MAC (V)~\cite{hudson2018compositional} and Object-based Attention~\cite{ding2020object} on the generalized CLEVRER dataset.
Tab.~\ref{tab:few-shot} shows our model outperforms other works by a large margin (75.6 vs. 26.7) on per question accuracy, demonstrating our model can quickly learn new concepts from few examples by reconstructing the physics world.
An example of generalization with few-shot data is shown in Fig.~\ref{fig:few-shot}. Our model learns a novel concept ``heavier'' from only 25 videos and the corresponding question-answer pairs. The simulation is then run with 5 times the mass to answer the question correctly.

\begin{table}[t]
	\begin{center}
	\small
	\setlength{\tabcolsep}{5pt}
	\resizebox{1\linewidth}{!}{%
	\begin{tabular}{lccccccccc}
	\toprule
        \multirow{2}{*}{Methods} & \multicolumn{2}{c}{Overall} & \multicolumn{2}{c}{Predictive} & \multicolumn{2}{c}{Counterfactual} & \multirow{2}{*}{Descriptive} & \multicolumn{2}{c}{Explanatory} \\ 
    \cmidrule(lr){2-3}\cmidrule(lr){4-5}\cmidrule(lr){6-7}\cmidrule(lr){9-10}
        & per task & per ques. & per opt. & per ques. & per opt. & per ques. &  & per opt. & per ques. \\ 
    \midrule
        Baseline & 72.6 & 81.6 & 85.1 & 72.4 & 77.6 & 49.6 & 87.8 & 88.0 & 80.6 \\
        + Collision-independent First & 81.3 & 87.8 & 86.1 & 72.8 & 89.3 & 74.1 & 91.3 & 91.9 & 86.9 \\
        + Curriculum Optimization & 85.6 & 90.2 & 87.6 & 76.5 & 94.8 & 84.3 & 92.2 & 93.3 & 89.2 \\
        + Re-optimization for Prediction & \textbf{90.3} & \textbf{92.0} & \textbf{95.7} & \textbf{91.4} & \textbf{94.8} & \textbf{84.3} & \textbf{93.4} & \textbf{96.3} & \textbf{91.9} \\
    \bottomrule
	\end{tabular}}
	\end{center}
	\caption{Ablation study on the optimization of physical parameters on CLEVRER~\cite{yi2019clevrer}. The reasoning accuracy for the four types of questions is continuously increased through a better learning process.
	}
	\label{tab:ablation_physics}
	\vspace{-10pt}
\end{table}

\noindent \textbf{Ablation Study on the Learning of Physics Models}~~
In this work, sample-independent physical parameters ($R$, $\lambda$) are learned from multiple training videos. In contrast, the sample-dependent parameters, such as $m, r, \alpha, v, l$, can only be learned with a single video, leading to difficulties in optimization, especially when there are many collisions.
This part studies the optimization of these sample-dependent parameters by making comparisons among the following four simplified learning processes on CLEVRER~\cite{yi2019clevrer}:
1) Baseline -- optimize all target parameters directly from all frames simultaneously.
2) Collision-independent First -- first use the frames before the collision to optimize collision-independent parameters for each object, including initial velocity $\overrightarrow{v_0}$, initial location $\overrightarrow{l_0}$, and initial angle $\alpha_0$; then optimize mass $m$ and restitution $r$ from all video frames.
3) Curriculum Optimization -- optimize $m$ and $r$ by performing multiple steps on [0, 40], [0, 80], and [0, 128] frames, where each step is initialized from the optimization of the previous step.
4) Re-optimization for Prediction (Full model) -- leverage the learned physical parameters as initialization and re-optimize all sample-dependent parameters with the last 20 frames to reduce the cumulative error over time.

Tab.~\ref{tab:ablation_physics} shows that the performance continuously increases when more optimization steps are used, demonstrating the contribution of each part. The ``Collision-independent First'' rule offers the greatest improvement, especially for counterfactual questions, as counterfactual simulations only rely on the initial state. ``Curriculum Optimization'' improves all types of questions, and ``Re-optimization for Prediction'' re-calculates the dynamics of the last 20 frames, thus mainly affect predictive questions.

\noindent \textbf{Failure Analysis}~~
\alias~learns the physics model from object trajectories in videos and language concepts in question-answer pairs. It is data-efficient and robust enough to work well when there exists inaccurate perception or incorrect concept learning in some video frames.
However, we noticed that the model might fail in the following cases:
1) If the object collides immediately after entering the image plane, there are insufficient frames before the collision to learn the initial velocity $v_0$.
2) If no collision occurs on an object, its restitution $r$ and mass $m$ cannot be optimized (unknown). We set default values for them.
3) The optimization becomes difficult if there are many cubes and collisions between them in the scene, because cube collisions (considering the sides and corners) are more complicated than sphere and cylinders'.
These issues are challenging and will be our future work.

\vspace{-5pt}
\begin{figure}[t]
\begin{minipage}[t]{0.50\linewidth}
\begin{figure}[H]
    \centering
    \includegraphics[width=\linewidth]{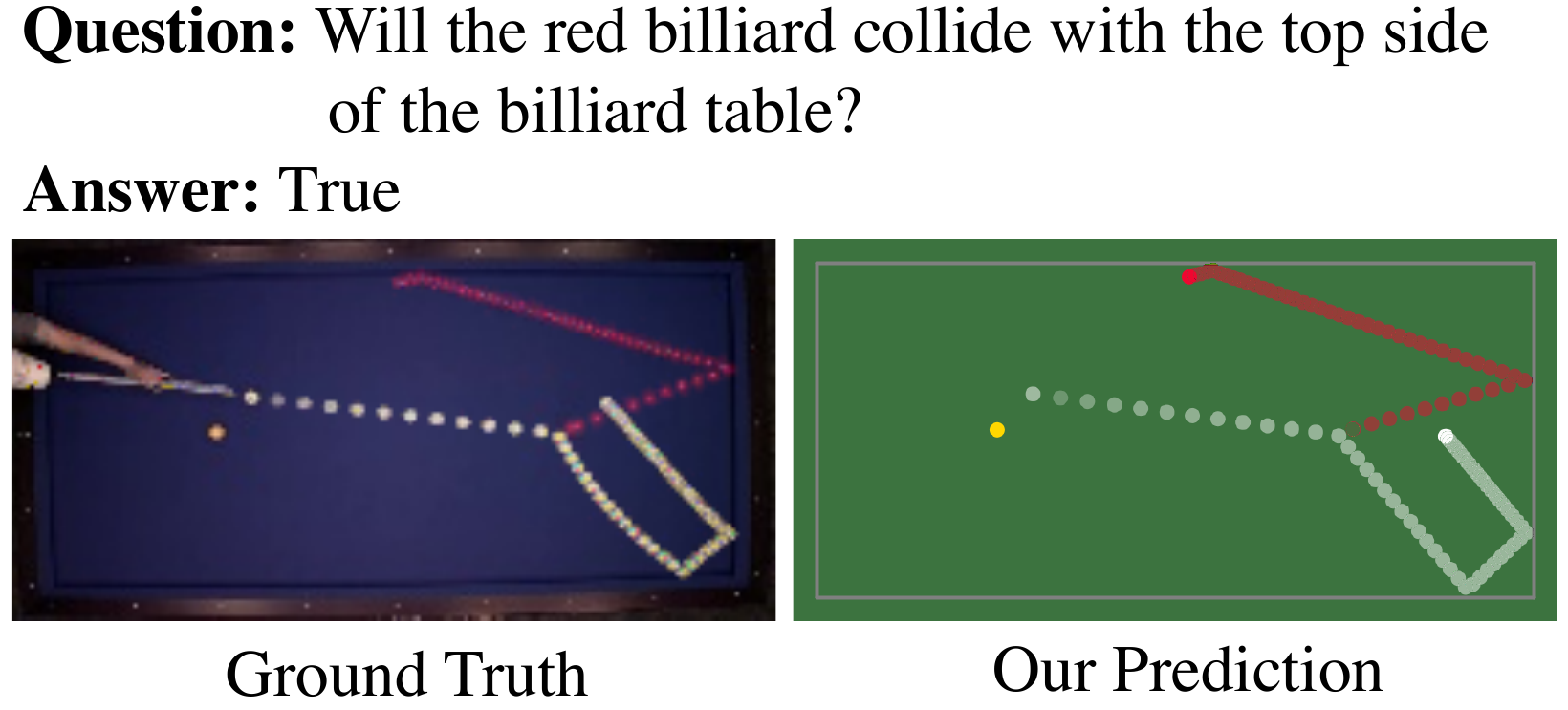}
    \vspace{-15pt}
    \caption{An example of physical simulation and question-answering on the real-world billiard dataset~\cite{qi2020learning}. \alias~learns accurate physics parameters and infers the correct answer by simulation.
    }
    \label{fig:billiards}
\end{figure}
\end{minipage}
\hfill
\begin{minipage}[t]{0.47\linewidth}
\begin{table}[H]
\centering
\setlength{\tabcolsep}{4pt}
\resizebox{1\linewidth}{!}{%
    \begin{tabular}{lcccc}
    \toprule
    \multirow{1}{*}{Methods}
    & S1$\text{ Err.}\downarrow$ 
    & S2$\text{ Err.}\downarrow$ 
    & QA Acc. (\%)$\uparrow$\\
    \midrule
    VIN~\cite{watters2017visual}
    & 1.02
    & 5.11 & 58.3 \\
    OM~\cite{janner2018reasoning}
    & 0.59
    & 3.23 & 61.1 \\
    CVP~\cite{ye2019compositional}
    & 3.57
    & 6.63 & 58.3 \\
    IN~\cite{battaglia2016interaction}
    & 0.37
    & 2.72 & 69.4 \\
    CIN~\cite{qi2020learning}
    & 0.30
    & 2.34 & 72.2 \\
    \midrule
    \alias~(Ours)
    & \textbf{0.24}
    & \textbf{0.88} & \textbf{80.6} \\
    \bottomrule
    \end{tabular}
}
\vspace{2pt}
\caption{Comparisons of the prediction error and question-answering accuracy on Real-Billiards. The rollout timesteps are chosen to be the same (S1) and twice (S2) as the training time ($T=20$). The error is scaled by 1,000. 
}
\label{tab:comparsion_billiards}
\end{table}
\end{minipage}
\vspace{-8pt}
\end{figure}

\subsection{Comparative Results on Real-World Billiards}
We also conduct experiments on the real-world dataset Real-billiard~\cite{qi2020learning} with our supplemented question-answer pairs.
Note that the billiard table is a chaotic system, and highly accurate long-term prediction is intractable.
Fig.~\ref{fig:billiards} shows an example of the ground truth video and our simulated prediction based on the perceptual grounded physics model. It can be seen that the predicted collision events and trajectories are of good quality.
Tab.~\ref{tab:comparsion_billiards} evaluates the prediction errors under two different rollout timesteps and QA accuracy with 5 competitors: VIN~\cite{watters2017visual}, OM~\cite{janner2018reasoning}, CVP~\cite{ye2019compositional}, IN~\cite{battaglia2016interaction}, and CIN~\cite{qi2020learning}.
For the prediction task, the rollout timesteps are chosen to be the same (S1$=[0, T]$) and twice (S2$=[T, 2T]$) as the training time, where the training time $T=20$.
We refer interested readers to CIN~\cite{qi2020learning} for more details. We find that \alias~is superior to these methods on both prediction and question answering tasks. Moreover, \alias~works well in long-term prediction. It reduces the S2 error on CIN~\cite{qi2020learning} by 62.4\%.

\section{Conclusion}
This work introduces \alias, a unified framework that integrates powerful differentiable physics models into dynamic visual reasoning. It contains three mutually beneficial components: a visual perception module, a concept learner, and a differentiable physics engine.
The visual perception module parses the input video into object trajectories and visual representations; the concept learner grounds language concepts and object attributes from question-answer pairs and the visual representations; with object trajectories and attributes as prior knowledge, the physics model optimizes all physical parameters of the scene and objects by differentiable simulation. With these explicit physical parameters, the physics model reruns the simulation to reason about future motion and causal events, which are then executed by a symbolic program executor to get the answer.
Equipped with the powerful physics model, \alias~is of highly data-efficient and generalizable that adapts to novel concepts quickly with few-shot data. Moreover, both the explicit physics engine and the symbolic executor are step-by-step interpretable, making \alias~fully transparent.
Extensive experiments on CLEVRER and Real-Billiards show that \alias~outperforms state-of-the-art dynamic reasoning methods by large margins.

\begin{ack}
This work was supported by MIT-IBM Watson AI Lab and its member company Nexplore, ONR MURI, DARPA Machine Common Sense program, ONR (N00014-18-1-2847), and Mitsubishi Electric; Ping Luo was supported by the General Research Fund of Hong Kong No.27208720.
\end{ack}

{\small
\bibliographystyle{abbrv}  
\bibliography{ref}

\begin{thebibliography}{10}

\bibitem{agrawal2016learning}
P.~Agrawal, A.~V. Nair, P.~Abbeel, J.~Malik, and S.~Levine.
\newblock Learning to poke by poking: Experiential learning of intuitive
  physics.
\newblock In {\em NeurIPS}, 2016.

\bibitem{ajay2019combining}
A.~Ajay, M.~Bauza, J.~Wu, N.~Fazeli, J.~B. Tenenbaum, A.~Rodriguez, and L.~P.
  Kaelbling.
\newblock Combining physical simulators and object-based networks for control.
\newblock In {\em ICRA}, 2019.

\bibitem{amizadeh2020neuro}
S.~Amizadeh, H.~Palangi, O.~Polozov, Y.~Huang, and K.~Koishida.
\newblock Neuro-symbolic visual reasoning: Disentangling" visual" from"
  reasoning".
\newblock In {\em ICML}, 2020.

\bibitem{andreas2016neural}
J.~Andreas, M.~Rohrbach, T.~Darrell, and D.~Klein.
\newblock Neural module networks.
\newblock In {\em CVPR}, 2016.

\bibitem{antol2015vqa}
S.~Antol, A.~Agrawal, J.~Lu, M.~Mitchell, D.~Batra, C.~Lawrence~Zitnick, and
  D.~Parikh.
\newblock Vqa: Visual question answering.
\newblock In {\em ICCV}, 2015.

\bibitem{bahdanau2014neural}
D.~Bahdanau, K.~Cho, and Y.~Bengio.
\newblock Neural machine translation by jointly learning to align and
  translate.
\newblock In {\em ICLR}, 2015.

\bibitem{bakhtin2019phyre}
A.~Bakhtin, L.~van~der Maaten, J.~Johnson, L.~Gustafson, and R.~Girshick.
\newblock Phyre: A new benchmark for physical reasoning.
\newblock In {\em NeurIPS}, 2019.

\bibitem{baradel2019cophy}
F.~Baradel, N.~Neverova, J.~Mille, G.~Mori, and C.~Wolf.
\newblock Cophy: Counterfactual learning of physical dynamics.
\newblock In {\em ICLR}, 2020.

\bibitem{battaglia2013simulation}
P.~W. Battaglia, J.~B. Hamrick, and J.~B. Tenenbaum.
\newblock Simulation as an engine of physical scene understanding.
\newblock {\em Proceedings of the National Academy of Sciences}, 2013.

\bibitem{battaglia2016interaction}
P.~W. Battaglia, R.~Pascanu, M.~Lai, D.~Rezende, and K.~Kavukcuoglu.
\newblock Interaction networks for learning about objects, relations and
  physics.
\newblock {\em arXiv preprint arXiv:1612.00222}, 2016.

\bibitem{bear2020learning}
D.~M. Bear, C.~Fan, D.~Mrowca, Y.~Li, S.~Alter, A.~Nayebi, J.~Schwartz,
  L.~Fei-Fei, J.~Wu, J.~B. Tenenbaum, et~al.
\newblock Learning physical graph representations from visual scenes.
\newblock In {\em NeurIPS}, 2020.

\bibitem{bengio2009curriculum}
Y.~Bengio, J.~Louradour, R.~Collobert, and J.~Weston.
\newblock Curriculum learning.
\newblock In {\em ICML}, pages 41--48, 2009.

\bibitem{butcher1975stability}
J.~C. Butcher.
\newblock A stability property of implicit runge-kutta methods.
\newblock {\em BIT Numerical Mathematics}, pages 358--361, 1975.

\bibitem{catto2009modeling}
E.~Catto.
\newblock Modeling and solving constraints.
\newblock In {\em Game Developers Conference}, page~16, 2009.

\bibitem{chang2016compositional}
M.~B. Chang, T.~Ullman, A.~Torralba, and J.~B. Tenenbaum.
\newblock A compositional object-based approach to learning physical dynamics.
\newblock In {\em ICLR}, 2017.

\bibitem{chen2021grounding}
Z.~Chen, J.~Mao, J.~Wu, K.-Y.~K. Wong, J.~B. Tenenbaum, and C.~Gan.
\newblock Grounding physical concepts of objects and events through dynamic
  visual reasoning.
\newblock In {\em ICLR}, 2021.

\bibitem{coumans2010bullet}
E.~Coumans.
\newblock Bullet physics engine.
\newblock {\em Open Source Software: http://bulletphysics. org}, 1(3):84, 2010.

\bibitem{de2018end}
F.~de~Avila Belbute-Peres, K.~Smith, K.~Allen, J.~Tenenbaum, and J.~Z. Kolter.
\newblock End-to-end differentiable physics for learning and control.
\newblock In {\em NeurIPS}, volume~31, pages 7178--7189, 2018.

\bibitem{degrave2019differentiable}
J.~Degrave, M.~Hermans, J.~Dambre, et~al.
\newblock A differentiable physics engine for deep learning in robotics.
\newblock {\em Frontiers in neurorobotics}, 13:6, 2019.

\bibitem{ding2020object}
D.~Ding, F.~Hill, A.~Santoro, and M.~Botvinick.
\newblock Object-based attention for spatio-temporal reasoning: Outperforming
  neuro-symbolic models with flexible distributed architectures.
\newblock {\em arXiv preprint arXiv:2012.08508}, 2020.

\bibitem{fan2019heterogeneous}
C.~Fan, X.~Zhang, S.~Zhang, W.~Wang, C.~Zhang, and H.~Huang.
\newblock Heterogeneous memory enhanced multimodal attention model for video
  question answering.
\newblock In {\em CVPR}, 2019.

\bibitem{finn2016unsupervised}
C.~Finn, I.~Goodfellow, and S.~Levine.
\newblock Unsupervised learning for physical interaction through video
  prediction.
\newblock In {\em NeurIPS}, 2016.

\bibitem{gan2017vqs}
C.~Gan, Y.~Li, H.~Li, C.~Sun, and B.~Gong.
\newblock {VQS}: Linking segmentations to questions and answers for supervised
  attention in vqa and question-focused semantic segmentation.
\newblock In {\em ICCV}, pages 1811--1820, 2017.

\bibitem{gan2020threedworld}
C.~Gan, J.~Schwartz, S.~Alter, M.~Schrimpf, J.~Traer, J.~De~Freitas,
  J.~Kubilius, A.~Bhandwaldar, N.~Haber, M.~Sano, et~al.
\newblock Threedworld: A platform for interactive multi-modal physical
  simulation.
\newblock {\em NeurIPS}, 2021.

\bibitem{gan2021transport}
C.~Gan, S.~Zhou, J.~Schwartz, S.~Alter, A.~Bhandwaldar, D.~Gutfreund, D.~L.
  Yamins, J.~J. DiCarlo, J.~McDermott, A.~Torralba, et~al.
\newblock The threedworld transport challenge: A visually guided
  task-and-motion planning benchmark for physically realistic embodied ai.
\newblock {\em arXiv preprint arXiv:2103.14025}, 2021.

\bibitem{girdhar2019cater}
R.~Girdhar and D.~Ramanan.
\newblock Cater: A diagnostic dataset for compositional actions and temporal
  reasoning.
\newblock In {\em ICLR}, 2020.

\bibitem{gkioxari2015finding}
G.~Gkioxari and J.~Malik.
\newblock Finding action tubes.
\newblock In {\em CVPR}, 2015.

\bibitem{han2019visual}
C.~Han, J.~Mao, C.~Gan, J.~Tenenbaum, and J.~Wu.
\newblock Visual concept-metaconcept learning.
\newblock In {\em NeurIPS}, 2019.

\bibitem{haninterpretable}
X.~Han, S.~Wang, C.~Su, W.~Zhang, Q.~Huang, and Q.~Tian.
\newblock Interpretable visual reasoning via probabilistic formulation under
  natural supervision.
\newblock In {\em ECCV}, 2020.

\bibitem{he2017mask}
K.~He, G.~Gkioxari, P.~Doll{\'a}r, and R.~Girshick.
\newblock Mask r-cnn.
\newblock In {\em ICCV}, pages 2961--2969, 2017.

\bibitem{he2016deep}
K.~He, X.~Zhang, S.~Ren, and J.~Sun.
\newblock Deep residual learning for image recognition.
\newblock In {\em CVPR}, 2016.

\bibitem{heiden2020neuralsim}
E.~Heiden, D.~Millard, E.~Coumans, Y.~Sheng, and G.~S. Sukhatme.
\newblock Neuralsim: Augmenting differentiable simulators with neural networks.
\newblock {\em arXiv preprint arXiv:2011.04217}, 2020.

\bibitem{heiden2019interactive}
E.~Heiden, D.~Millard, H.~Zhang, and G.~S. Sukhatme.
\newblock Interactive differentiable simulation.
\newblock {\em arXiv preprint arXiv:1905.10706}, 2019.

\bibitem{HochSchm97}
S.~Hochreiter and J.~Schmidhuber.
\newblock Long short-term memory.
\newblock {\em Neural Computation}, 9(8):1735--1780, 1997.

\bibitem{hu2018explainable}
R.~Hu, J.~Andreas, T.~Darrell, and K.~Saenko.
\newblock Explainable neural computation via stack neural module networks.
\newblock In {\em ECCV}, 2018.

\bibitem{hu2019difftaichi}
Y.~Hu, L.~Anderson, T.-M. Li, Q.~Sun, N.~Carr, J.~Ragan-Kelley, and F.~Durand.
\newblock Difftaichi: Differentiable programming for physical simulation.
\newblock In {\em ICLR}, 2020.

\bibitem{huang2020location}
D.~Huang, P.~Chen, R.~Zeng, Q.~Du, M.~Tan, and C.~Gan.
\newblock Location-aware graph convolutional networks for video question
  answering.
\newblock In {\em AAAI}, pages 11021--11028, 2020.

\bibitem{huang2021plasticinelab}
Z.~Huang, Y.~Hu, T.~Du, S.~Zhou, H.~Su, J.~B. Tenenbaum, and C.~Gan.
\newblock Plasticinelab: A soft-body manipulation benchmark with differentiable
  physics.
\newblock In {\em ICLR}, 2021.

\bibitem{hudson2018compositional}
D.~A. Hudson and C.~D. Manning.
\newblock Compositional attention networks for machine reasoning.
\newblock In {\em ICLR}, 2018.

\bibitem{hudson2019gqa}
D.~A. Hudson and C.~D. Manning.
\newblock Gqa: A new dataset for real-world visual reasoning and compositional
  question answering.
\newblock In {\em CVPR}, 2019.

\bibitem{jang2017tgif}
Y.~Jang, Y.~Song, Y.~Yu, Y.~Kim, and G.~Kim.
\newblock Tgif-qa: Toward spatio-temporal reasoning in visual question
  answering.
\newblock In {\em CVPR}, 2017.

\bibitem{janner2018reasoning}
M.~Janner, S.~Levine, W.~T. Freeman, J.~B. Tenenbaum, C.~Finn, and J.~Wu.
\newblock Reasoning about physical interactions with object-oriented prediction
  and planning.
\newblock In {\em ICLR}, 2019.

\bibitem{jatavallabhula2021gradsim}
K.~M. Jatavallabhula, M.~Macklin, F.~Golemo, V.~Voleti, L.~Petrini, M.~Weiss,
  B.~Considine, J.~Parent-Levesque, K.~Xie, K.~Erleben, et~al.
\newblock gradsim: Differentiable simulation for system identification and
  visuomotor control.
\newblock In {\em ICLR}, 2021.

\bibitem{johnson2017clevr}
J.~Johnson, B.~Hariharan, L.~van~der Maaten, L.~Fei-Fei, C.~Lawrence~Zitnick,
  and R.~Girshick.
\newblock Clevr: A diagnostic dataset for compositional language and elementary
  visual reasoning.
\newblock In {\em CVPR}, 2017.

\bibitem{johnson2017inferring}
J.~Johnson, B.~Hariharan, L.~Van Der~Maaten, J.~Hoffman, L.~Fei-Fei,
  C.~Lawrence~Zitnick, and R.~Girshick.
\newblock Inferring and executing programs for visual reasoning.
\newblock In {\em ICCV}, pages 2989--2998, 2017.

\bibitem{kipf2019contrastive}
T.~Kipf, E.~van~der Pol, and M.~Welling.
\newblock Contrastive learning of structured world models.
\newblock In {\em ICLR}, 2020.

\bibitem{kipf2017semi}
T.~N. Kipf and M.~Welling.
\newblock Semi-supervised classification with graph convolutional networks.
\newblock In {\em ICLR}, 2017.

\bibitem{le2020hierarchical}
T.~M. Le, V.~Le, S.~Venkatesh, and T.~Tran.
\newblock Hierarchical conditional relation networks for video question
  answering.
\newblock In {\em CVPR}, pages 9972--9981, 2020.

\bibitem{lei2018tvqa}
J.~Lei, L.~Yu, M.~Bansal, and T.~L. Berg.
\newblock Tvqa: Localized, compositional video question answering.
\newblock In {\em EMNLP}, 2018.

\bibitem{lerer2016learning}
A.~Lerer, S.~Gross, and R.~Fergus.
\newblock Learning physical intuition of block towers by example.
\newblock In {\em ICML}, 2016.

\bibitem{li2019beyond}
X.~Li, J.~Song, L.~Gao, X.~Liu, W.~Huang, X.~He, and C.~Gan.
\newblock Beyond rnns: Positional self-attention with co-attention for video
  question answering.
\newblock In {\em AAAI}, pages 8658--8665, 2019.

\bibitem{li2020visual}
Y.~Li, T.~Lin, K.~Yi, D.~Bear, D.~Yamins, J.~Wu, J.~Tenenbaum, and A.~Torralba.
\newblock Visual grounding of learned physical models.
\newblock In {\em ICML}, pages 5927--5936, 2020.

\bibitem{li2018learning}
Y.~Li, J.~Wu, R.~Tedrake, J.~B. Tenenbaum, and A.~Torralba.
\newblock Learning particle dynamics for manipulating rigid bodies, deformable
  objects, and fluids.
\newblock In {\em ICLR}, 2019.

\bibitem{li2019propagation}
Y.~Li, J.~Wu, J.-Y. Zhu, J.~B. Tenenbaum, A.~Torralba, and R.~Tedrake.
\newblock Propagation networks for model-based control under partial
  observation.
\newblock In {\em ICRA}, 2019.

\bibitem{liang2020differentiable}
J.~Liang and M.~C. Lin.
\newblock Differentiable physics simulation.
\newblock In {\em ICLR Workshop}, 2020.

\bibitem{lin2017feature}
T.-Y. Lin, P.~Doll{\'a}r, R.~Girshick, K.~He, B.~Hariharan, and S.~Belongie.
\newblock Feature pyramid networks for object detection.
\newblock In {\em CVPR}, pages 2117--2125, 2017.

\bibitem{lin2014microsoft}
T.-Y. Lin, M.~Maire, S.~Belongie, J.~Hays, P.~Perona, D.~Ramanan,
  P.~Doll{\'a}r, and C.~L. Zitnick.
\newblock Microsoft coco: Common objects in context.
\newblock In {\em ECCV}, pages 740--755. Springer, 2014.

\bibitem{liu1989limited}
D.~C. Liu and J.~Nocedal.
\newblock On the limited memory bfgs method for large scale optimization.
\newblock {\em Mathematical programming}, pages 503--528, 1989.

\bibitem{locatello2020object}
F.~Locatello, D.~Weissenborn, T.~Unterthiner, A.~Mahendran, G.~Heigold,
  J.~Uszkoreit, A.~Dosovitskiy, and T.~Kipf.
\newblock Object-centric learning with slot attention.
\newblock {\em arXiv preprint arXiv:2006.15055}, 2020.

\bibitem{mao2019neuro}
J.~Mao, C.~Gan, P.~Kohli, J.~B. Tenenbaum, and J.~Wu.
\newblock The neuro-symbolic concept learner: Interpreting scenes, words, and
  sentences from natural supervision.
\newblock In {\em ICLR}, 2019.

\bibitem{mascharka2018transparency}
D.~Mascharka, P.~Tran, R.~Soklaski, and A.~Majumdar.
\newblock Transparency by design: Closing the gap between performance and
  interpretability in visual reasoning.
\newblock In {\em CVPR}, pages 4942--4950, 2018.

\bibitem{misra2018learning}
I.~Misra, R.~Girshick, R.~Fergus, M.~Hebert, A.~Gupta, and L.~Van Der~Maaten.
\newblock Learning by asking questions.
\newblock In {\em CVPR}, pages 11--20, 2018.

\bibitem{mottaghi2016happens}
R.~Mottaghi, M.~Rastegari, A.~Gupta, and A.~Farhadi.
\newblock “what happens if...” learning to predict the effect of forces in
  images.
\newblock In {\em ECCV}. Springer, 2016.

\bibitem{muller2008real}
M.~M{\"u}ller, J.~Stam, D.~James, and N.~Th{\"u}rey.
\newblock Real time physics: class notes.
\newblock In {\em ACM SIGGRAPH 2008 classes}, pages 1--90, 2008.

\bibitem{qi2020learning}
H.~Qi, X.~Wang, D.~Pathak, Y.~Ma, and J.~Malik.
\newblock Learning long-term visual dynamics with region proposal interaction
  networks.
\newblock In {\em ICLR}, 2021.

\bibitem{riochet2018intphys}
R.~Riochet, M.~Y. Castro, M.~Bernard, A.~Lerer, R.~Fergus, V.~Izard, and
  E.~Dupoux.
\newblock Intphys: A framework and benchmark for visual intuitive physics
  reasoning.
\newblock {\em arXiv preprint arXiv:1803.07616}, 2018.

\bibitem{shao2014imagining}
T.~Shao, A.~Monszpart, Y.~Zheng, B.~Koo, W.~Xu, K.~Zhou, and N.~J. Mitra.
\newblock Imagining the unseen: Stability-based cuboid arrangements for scene
  understanding.
\newblock {\em ACM TOG}, 2014.

\bibitem{smith2019modeling}
K.~Smith, L.~Mei, S.~Yao, J.~Wu, E.~Spelke, J.~Tenenbaum, and T.~Ullman.
\newblock Modeling expectation violation in intuitive physics with coarse
  probabilistic object representations.
\newblock {\em NeurIPS}, 32:8985--8995, 2019.

\bibitem{MovieQA}
M.~Tapaswi, Y.~Zhu, R.~Stiefelhagen, A.~Torralba, R.~Urtasun, and S.~Fidler.
\newblock {MovieQA: Understanding Stories in Movies through
  Question-Answering}.
\newblock In {\em CVPR}, 2016.

\bibitem{toussaint2018differentiable}
M.~A. Toussaint, K.~R. Allen, K.~A. Smith, and J.~B. Tenenbaum.
\newblock Differentiable physics and stable modes for tool-use and manipulation
  planning.
\newblock In {\em IJCAI}, 2019.

\bibitem{vaswani2017attention}
A.~Vaswani, N.~Shazeer, N.~Parmar, J.~Uszkoreit, L.~Jones, A.~N. Gomez,
  {\L}.~Kaiser, and I.~Polosukhin.
\newblock Attention is all you need.
\newblock In {\em NeurIPS}, pages 5998--6008, 2017.

\bibitem{veerapaneni2020entity}
R.~Veerapaneni, J.~D. Co-Reyes, M.~Chang, M.~Janner, C.~Finn, J.~Wu,
  J.~Tenenbaum, and S.~Levine.
\newblock Entity abstraction in visual model-based reinforcement learning.
\newblock In {\em CoRL}, pages 1439--1456, 2020.

\bibitem{watters2017visual}
N.~Watters, D.~Zoran, T.~Weber, P.~Battaglia, R.~Pascanu, and A.~Tacchetti.
\newblock Visual interaction networks: Learning a physics simulator from video.
\newblock {\em NeurIPS}, 30:4539--4547, 2017.

\bibitem{wu2017learning}
J.~Wu, E.~Lu, P.~Kohli, B.~Freeman, and J.~Tenenbaum.
\newblock Learning to see physics via visual de-animation.
\newblock In {\em NeurIPS}, pages 153--164, 2017.

\bibitem{galileo}
J.~Wu, I.~Yildirim, J.~J. Lim, W.~T. Freeman, and J.~B. Tenenbaum.
\newblock Galileo: Perceiving physical object properties by integrating a
  physics engine with deep learning.
\newblock In {\em NeurIPS}, 2015.

\bibitem{Wu_2016_CVPR}
Q.~Wu, C.~Shen, L.~Liu, A.~Dick, and A.~van~den Hengel.
\newblock What value do explicit high level concepts have in vision to language
  problems?
\newblock In {\em CVPR}, 2016.

\bibitem{xu2017video}
D.~Xu, Z.~Zhao, J.~Xiao, F.~Wu, H.~Zhang, X.~He, and Y.~Zhuang.
\newblock Video question answering via gradually refined attention over
  appearance and motion.
\newblock In {\em ACM MM}, 2017.

\bibitem{ye2018interpretable}
T.~Ye, X.~Wang, J.~Davidson, and A.~Gupta.
\newblock Interpretable intuitive physics model.
\newblock In {\em ECCV}, 2018.

\bibitem{ye2019compositional}
Y.~Ye, M.~Singh, A.~Gupta, and S.~Tulsiani.
\newblock Compositional video prediction.
\newblock In {\em ICCV}, pages 10353--10362, 2019.

\bibitem{ye2017video}
Y.~Ye, Z.~Zhao, Y.~Li, L.~Chen, J.~Xiao, and Y.~Zhuang.
\newblock Video question answering via attribute-augmented attention network
  learning.
\newblock In {\em ICLR}, 2017.

\bibitem{yi2019clevrer}
K.~Yi, C.~Gan, Y.~Li, P.~Kohli, J.~Wu, A.~Torralba, and J.~B. Tenenbaum.
\newblock Clevrer: Collision events for video representation and reasoning.
\newblock In {\em ICLR}, 2020.

\bibitem{yi2018neural}
K.~Yi, J.~Wu, C.~Gan, A.~Torralba, P.~Kohli, and J.~Tenenbaum.
\newblock Neural-symbolic vqa: Disentangling reasoning from vision and language
  understanding.
\newblock In {\em NeurIPS}, 2018.

\bibitem{zadeh2019social}
A.~Zadeh, M.~Chan, P.~P. Liang, E.~Tong, and L.-P. Morency.
\newblock Social-iq: A question answering benchmark for artificial social
  intelligence.
\newblock In {\em CVPR}, 2019.

\bibitem{zhou2021hopper}
H.~Zhou, A.~Kadav, F.~Lai, A.~Niculescu-Mizil, M.~R. Min, M.~Kapadia, and H.~P.
  Graf.
\newblock Hopper: Multi-hop transformer for spatiotemporal reasoning.
\newblock In {\em ICLR}, 2021.

\bibitem{zhu2016visual7w}
Y.~Zhu, O.~Groth, M.~Bernstein, and L.~Fei-Fei.
\newblock Visual7w: Grounded question answering in images.
\newblock In {\em CVPR}, 2016.

\end{thebibliography}
}

\newpage



\appendix
\setcounter{section}{0}
\begin{appendices}

\section{Appendix}
In this section, we provide supplementary details of our \alias.
First, we give more details of our physics model and the neuro-symbolic operations in the program executor.
We then introduce the datasets we use and build, including a synthetic dataset (CLEVRER~\cite{yi2019clevrer}), a real-world dataset (Real-Billiard~\cite{qi2020learning}), and a newly built few-shot dataset (Generalized CLEVRER).
After that, we detail the training settings and steps.

\subsection{Details of Physics Model}
In this part, we provide supplementary details of our physics model.
With the perceptually grounded object shapes and trajectories from the perception module and the concept learner of \alias, our physics model performs differentiable simulation to optimize the physical parameters of the scene and objects by comparing the simulation $L'$ with the video observations $L^\text{BEV}$. The target bird's-eye view (BEV) trajectory $L^\text{BEV}$ is obtained by projecting the object center to the BEV coordinate.
The Camera-to-BEV projection can be written as:
\begin{equation}
\begin{bmatrix}
    x \\
    y \\
    - \\
    1 \\
  \end{bmatrix}_{BEV}  = \mathbf{K}^{-1} \cdot \begin{bmatrix}
    ~x \cdot z~  \\
    ~y \cdot z~  \\
    ~z~ \\
    ~1~ \\
  \end{bmatrix}_{camera} 
\label{eqn:proj}
\end{equation}
where $\mathbf{K}$ is the estimated camera matrix, $[x,y,z]_{camera}$ is the point in 2D image coordinates ($z_{camera}$ can be calculated from the camera matrix $\mathbf{K}$), $[x,y]_{BEV}$ denotes the horizontal position and vertical position of the projected point in BEV coordinates.

Based on the graphics programming language DiffTaichi~\cite{hu2019difftaichi}, our physics model is implemented as \engine. Based on conservation of momentum and angular momentum, it iteratively simulates a small time step of $\Delta t$ based on the objects' state in the BEV coordinate through inferring collision events, forces and impulses acting on the object, and updating the state of each object.
In addition to calculating the acceleration based on the conservation of momentum in our main paper, we also calculate the angular acceleration based on the angular momentum. For example, we have: $\overrightarrow{M}=\overrightarrow{r} \times \overrightarrow{F}$ and $M=I \frac{\mathrm{d}\omega}{\mathrm{d}t}$, where $\overrightarrow{M}$ denotes moment of force, $\overrightarrow{F}$ is the applied force, and $\overrightarrow{r}$ is the distance from the applied force to object. The momentum of inertia $I$ is $1/6m(2R)^2$ for the cube, where $m$ represents its mass.

\begin{figure}[htbp]
    \centering
    \begin{minipage}[c]{.36\textwidth} 
    \includegraphics[width=1\linewidth]{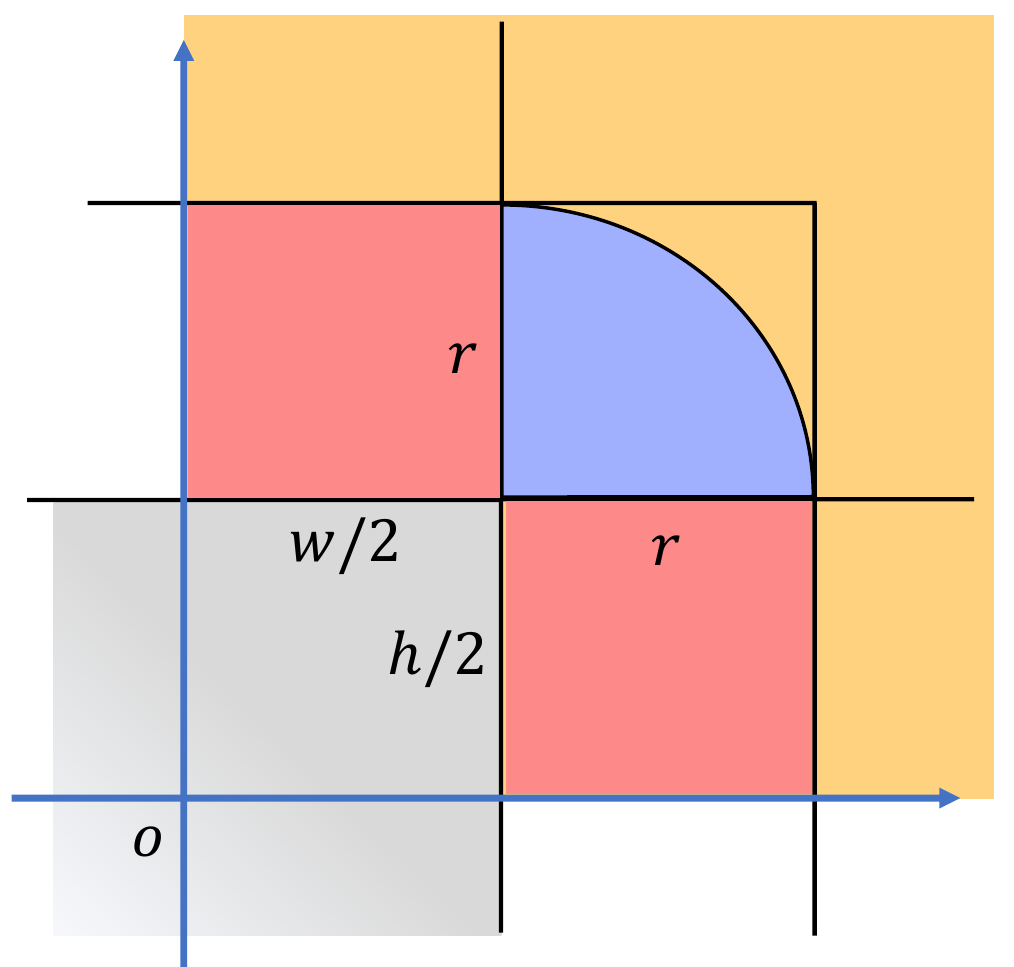}
    \end{minipage}
    ~~~~
    \begin{minipage}[c]{.58\textwidth}
    \caption{An illustration of circle-rectangle collision detection. 
    The gray part denotes the rectangle (cube) and we transform the origin to the center of the rectangle so that the coordinate axis is parallel to its side.
    For each simulation step: we consider three situations: 1) if the center of the circle is in the orange area, the circle and the rectangle do not collide;
    2) if the circle center is in the red area, the circle collides with the rectangle and the collision direction is perpendicular to the coordinate axis; 
    3) if the circle center falls in the purple area, the circle and the rectangle collide and the collision direction is perpendicular to the tangent of the collision position on the circle.
    }
    \label{fig:collision}
    \end{minipage}
\end{figure}

In this work, we perform collision detection between circles and rectangles in BEV view. Fig.~\ref{fig:collision} shows the illustration of our circle-rectangle collision detection algorithm.
We project the center of the rectangle (the gray part) to the origin so that the coordinate axis is parallel to its side. 
Then the area outside the rectangle is divided into three parts that the center of the circle can fall: 1) collision with the sides of the square (red); 2) collision with the corners of the square (purple); 3) no collision (orange).
The implementation of circle-circle and rectangle-rectangle collisions is similar.

\begin{figure}[t]
    \centering
    \includegraphics[width=\linewidth]{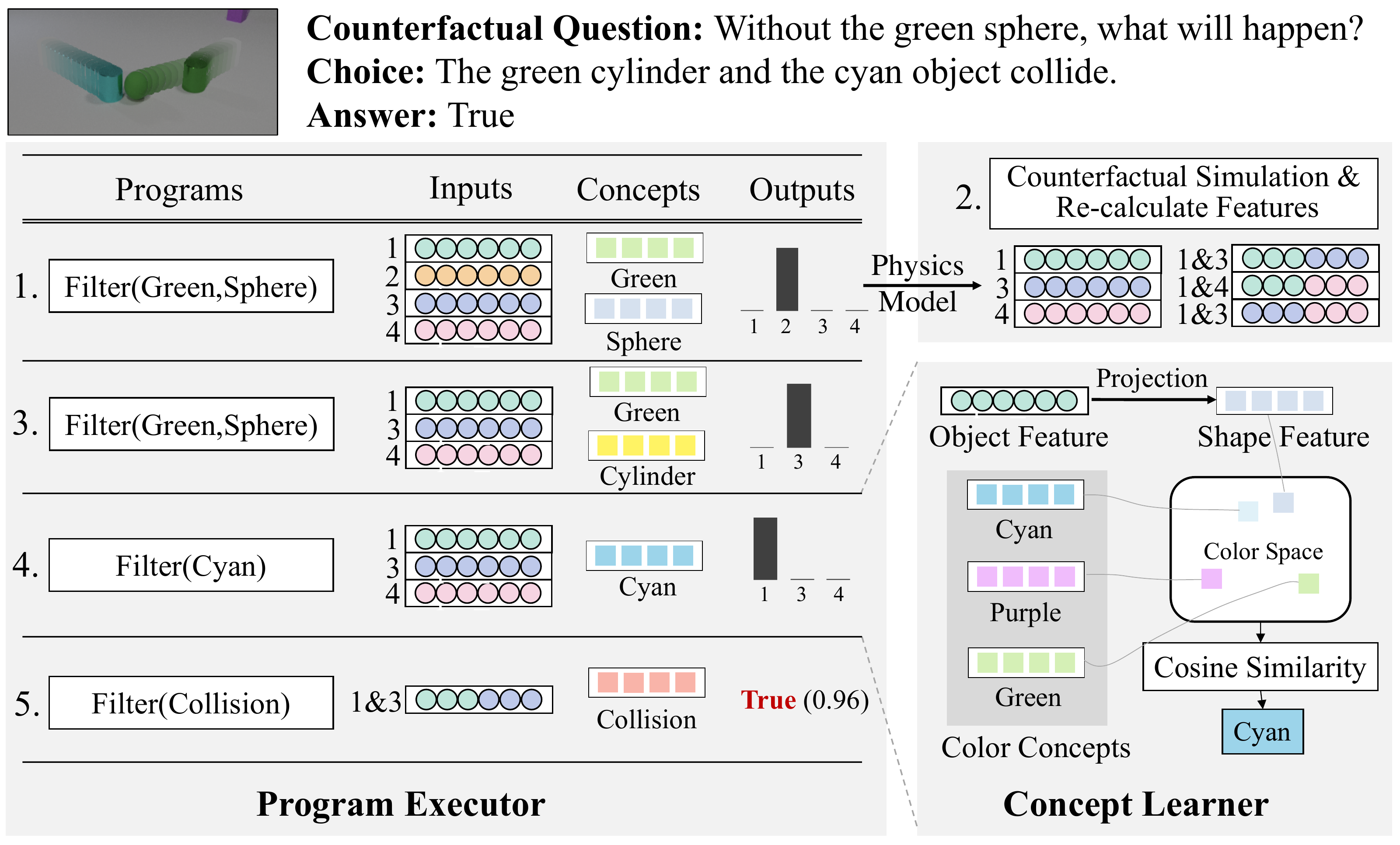}
    \caption{An illustration of the reasoning process of the program executor and concept learner. The program executor executes the parsed programs (\eg, Filter\_static\_concept (color, shape, material)) step-by-step with the visual representations and language concepts. For each step, it leverages the concept learner or physical model to filter specific targets or simulate/predict new visual trajectories.}
    \label{fig:illustration}
\end{figure}

\subsection{Details of Neuro-Symbolic Programs}
Following DCL~\cite{chen2021grounding}, we represent the objects, events and moments through learnable embeddings and quantize the static and dynamic concepts to perform temporal and causal reasoning. In this part, we list all the available data types and operations for CLEVRER in Tab.~\ref{tb:operation}. 
We refer interested readers to DCL~\cite{chen2021grounding} for more details.

We also visualize the reasoning process of an example step-by-step in Fig.~\ref{fig:illustration}. It shows how we get the correct answer for the counterfactual question `Without the green sphere, what will happen?' with a choice `The green cylinder and the cyan object collide'. After the first program `Filter\_static\_concept(all objects, green sphere)' is executed, the executor removes the retrieved object, reruns the simulation to get counterfactual trajectories, and updates the visual features. After that, the executor runs the remaining programs and gets the final answer `True' with a probability of 0.96 calculated through the cosine distance in the concept learner.

\subsection{Details of Datasets}
\noindent \textbf{CLEVRER}~
CLEVRER~\cite{yi2019clevrer} is a diagnostic video dataset for systematic evaluation of computational models on a wide range of reasoning tasks. 
Objects in CLEVRER videos adopt similar compositional intrinsic attributes as in CLEVR~\cite{johnson2017clevr}, including three shapes (cube, sphere, and cylinder), two materials (metal and rubber), and eight colours (gray, red, blue, green, brown, cyan, purple, and yellow). All objects have the same size, same friction coefficient (except the sphere that rolling on the ground), so no vertical bouncing occurs during the collision. Each object has a different mass and a different restitution coefficient. CLEVRER introduces three types of events: enter, exit and collision, each of which contains a fixed number of object participants: 2 for collision and 1 for enter and exit. The objects and events form an abstract representation of the video.

CLEVRER includes four types of question: descriptive (\eg `what color'), explanatory (`what’s responsible for'), predictive (`what will happen next'), and counterfactual (`what if'), where the first two types concern more on video understanding and temporal reasoning, while the latter two types involve physical dynamics and predictions in reasoning. Therefore, we mainly focus on the predictive and counterfactual questions in this work. 
CLEVRER consists of 2,000 videos, with a number of 1,000 training videos, 5,000 validation videos, and 5,000 test videos. It also contains 219,918 descriptive questions, 33,811 explanatory questions, 14,298 predictive questions, and 37,253 counterfactual questions. In this paper, we tune the model using the validation set and evaluate it with the test set. 

\noindent \textbf{Generalized CLEVRER}~
To evaluate the generalizability of reasoning methods, we collect a few-shot physical reasoning dataset with novel language and physical concepts (\eg, `heavier' and `lighter'), termed generalized CLEVRER, containing 100 videos (split into 25/25/50 for train/validation/test) with 375 options in 158 counterfactual questions. This dataset is supplementary to CLEVRER~\cite{yi2019clevrer} for generalizing to new concepts (\ie, heavier, lighter) with very few samples. All videos last for 5 seconds and are generated by a physics engine~\cite{coumans2010bullet} that simulates object motion plus a graphs engine that renders the frames. It has the same video settings (objects and events settings) with CLEVRER but different questions/concepts, \eg, ``What would happen if the blue sphere were heavier?'', we generate the ground truth video in the counterfactual case by setting five times the weight and perform the physical simulation with Bullet~\cite{coumans2010bullet}. In this work, we evaluate the QA accuracy of this dataset.

\noindent \textbf{Real-Billiard}~
For real-world scenarios, we conduct experiments on the Real-Billiard~\cite{qi2020learning} dataset, which contains three-cushion billiards videos captured in real games for dynamics prediction. There are 62 training videos with 18,306 frames, and 5 testing videos with 1,995 frames.
The bounding box annotations are from an off-the-shelf ResNet-101 FPN detector~\cite{lin2017feature} pretrained on COCO~\cite{lin2014microsoft} and fine-tuned on a subset of 30 images from our dataset. Wrong detections are manually filtered out.
We generate 6 reasoning questions (\eg, ``will one billiard collide with ...?'') for each video and evaluate both the prediction error and QA accuracy.

\subsection{Details of Training Settings}
As in \cite{yi2019clevrer,chen2021grounding}, we use a pre-trained Faster R-CNN model~\cite{he2017mask} that is trained on 4,000 video frames randomly sampled from the training set with object masks and attribute annotations to generate object proposals for each frame. We train the language program parser with 1,000 programs for all question types.
All deep modules (concept learner and program executor) are trained using Adam optimizer for 40 epochs on 8 Nvidia 1080Ti GPUs and the learning rate is set to $10^{-4}$. The camera matrix is optimized from 20 training videos. We set $\Delta t=0.004\text{s}, D=256, C=64, K=10, S=10$, and $T=128$ for CLEVRER~\cite{yi2019clevrer} and $T=20$ for Real-Billiard~\cite{qi2020learning}. In addition to our standard model that grounds object properties from question-answer pairs, we also train a variant (\alias~$\dag$) on CLEVRER with an explicit rule-based program executor~\cite{yi2019clevrer} and object attribute supervisions (attribute annotation in 4000 frames learned by the Faster R-CNN model). 

For the physical model, we use the L-BFGS optimizer~\cite{liu1989limited} with an adaptive learning rate to optimize all physical parameters. 
The optimization terminates when it reaches a certain number of steps or the loss is less than a certain value. In all experiments, the number of the optimization step is set to 20. The loss threshold is set to 0.0005 for the learning of collision-independent parameters (\ie, initial velocity, initial location, and initial angle), and 0.0002, 0.001, 0.01 for the optimization of collision-dependent parameters (mass and restitution) on [0, 40], [0, 80], and [0, 128] frames, respectively.

The training of \alias~can be summarized into three stages. First, we extract the visual features directly from the video by the visual perception module, and learn language concepts in the concept learner from all descriptive and explanatory questions;
second, we optimize all physical parameters by using the perceived trajectories and the learned concepts;
third, after obtaining the physics model, we re-calculate the visual features from the simulated trajectories and finetune language concept embeddings from all question types, including predictive and counterfactual questions.
During this training process, the three parts of \alias~are integrated seamlessly and benefit each other.

\subsection{Visualizations}
\begin{figure}[htbp!]
    \centering
    \includegraphics[width=\linewidth]{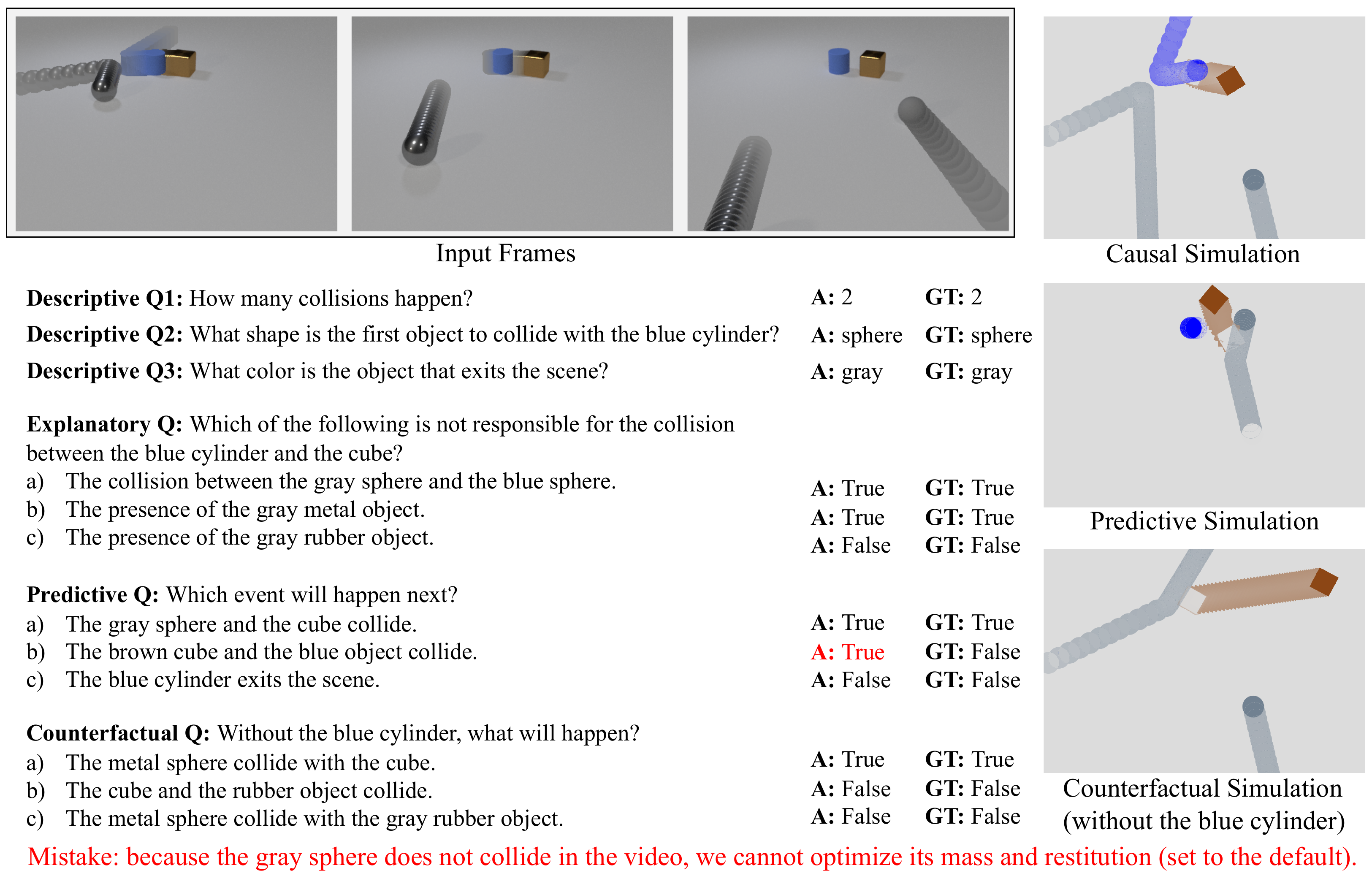}
----------------------------------------------------------------------------------------------------------------------\\~~\\~~
    \includegraphics[width=\linewidth]{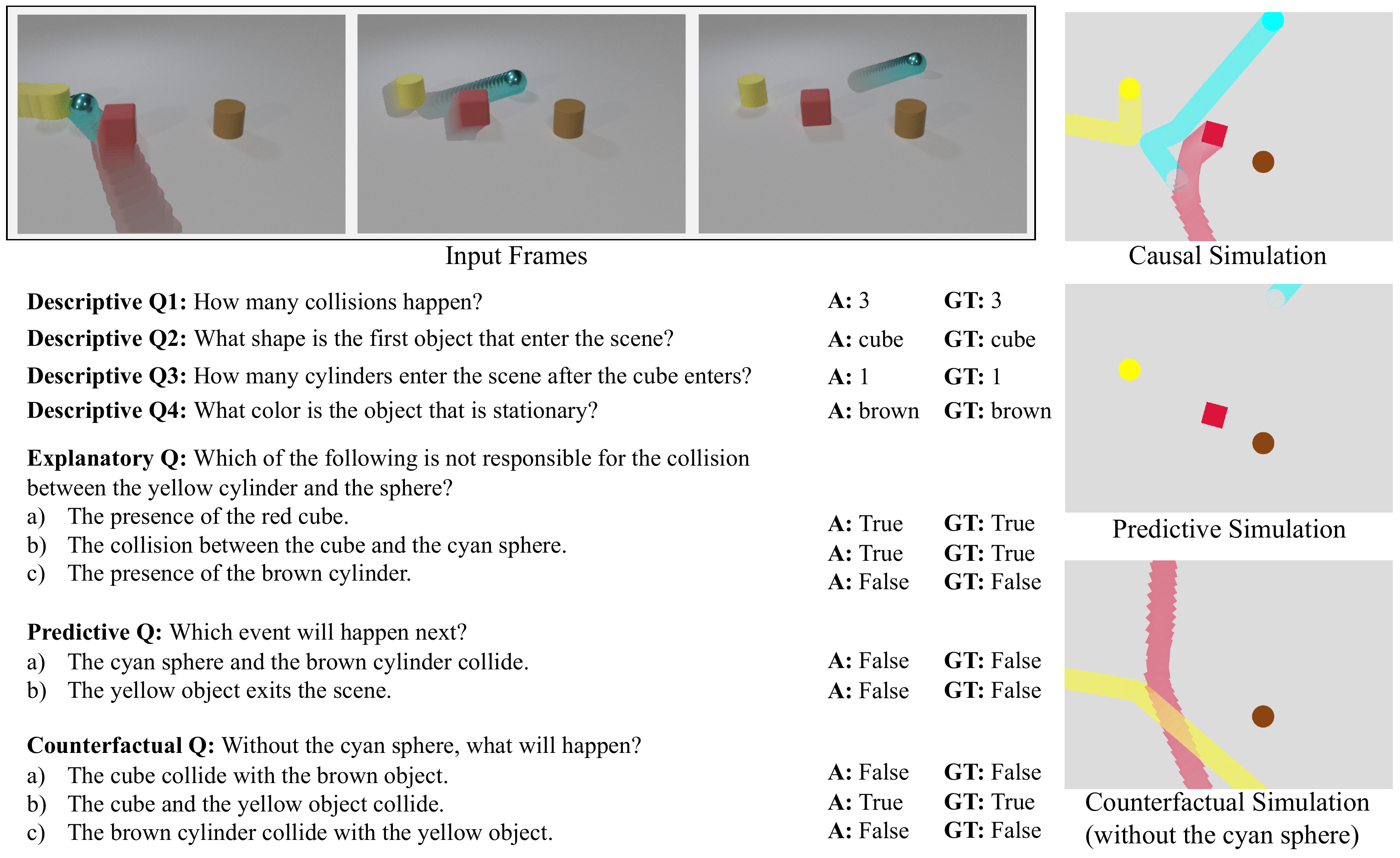}
    \caption{Visualization (1) of the videos and question-answering results of our \alias~on CLEVRER.
    We highlighted our failure in \textcolor{red}{red} and explained the cause of it.}
    \label{fig:examples_0}
\end{figure}

\begin{figure}[htbp!]
    \centering
    \includegraphics[width=\linewidth]{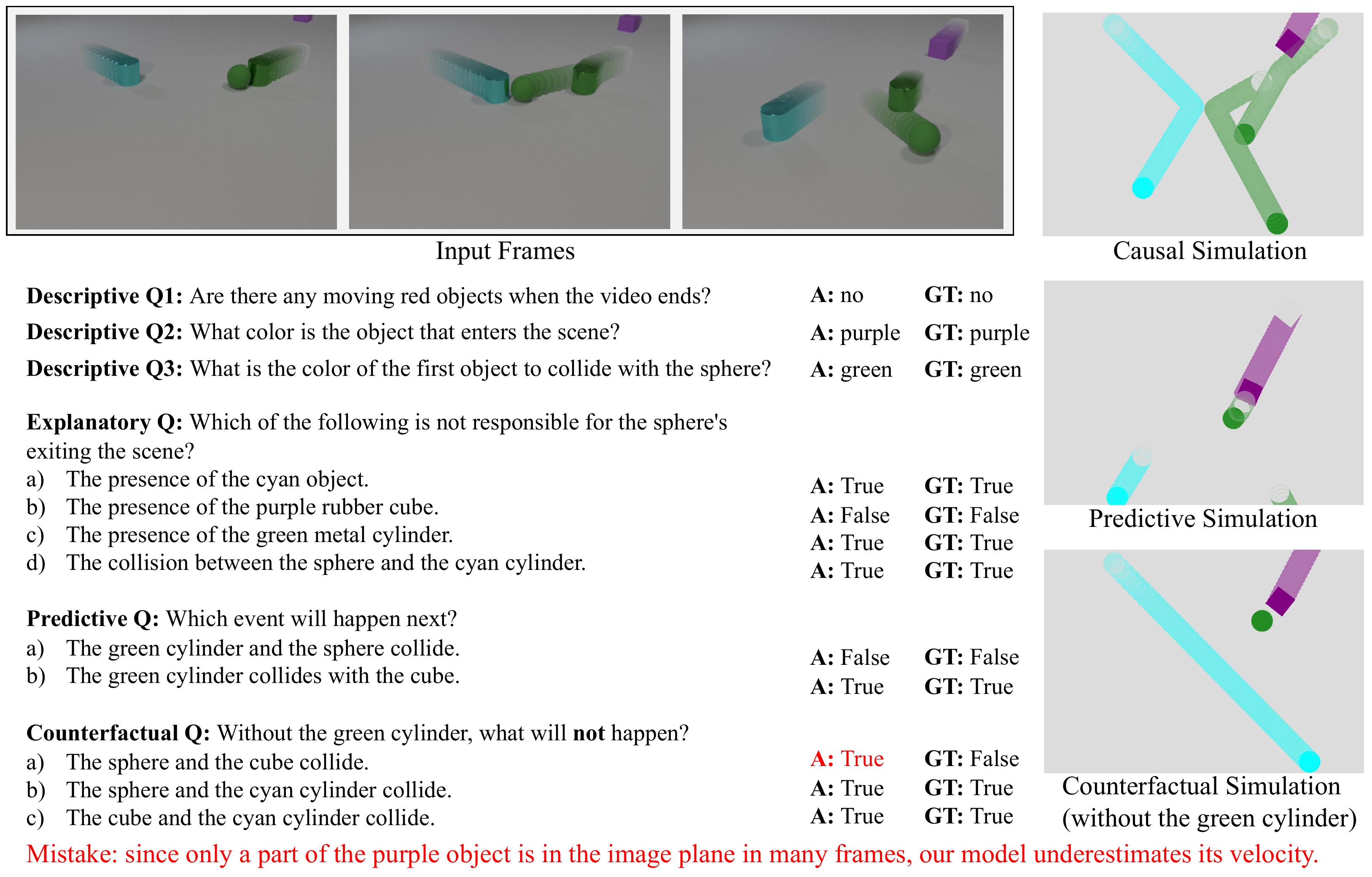}
----------------------------------------------------------------------------------------------------------------------\\~~\\~~
    \includegraphics[width=\linewidth]{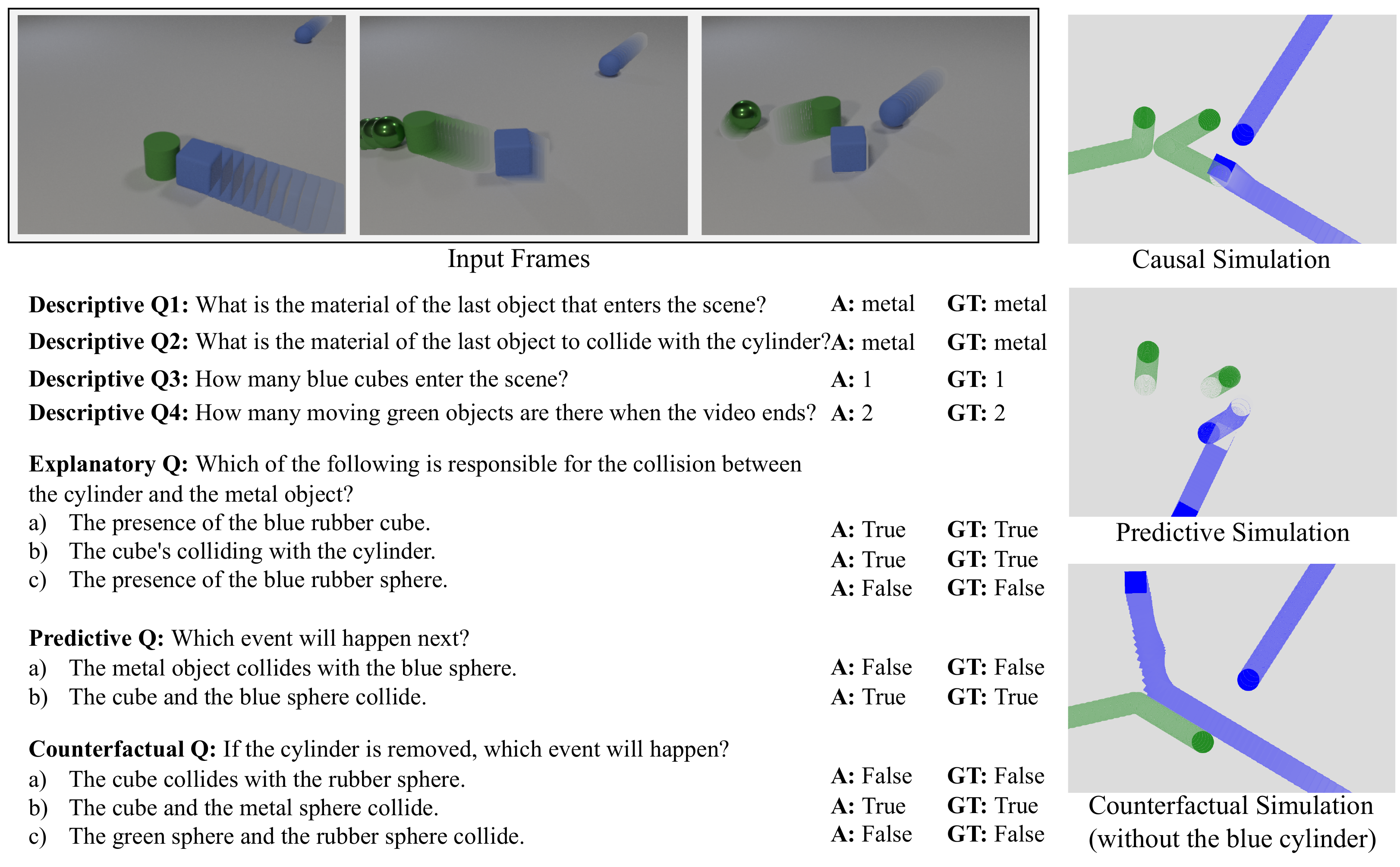}
    \caption{Visualization (2) of the videos and question-answering results of our \alias~on CLEVRER. We highlighted our failure in \textcolor{red}{red} and explained the cause of it.}
    \label{fig:examples_1}
\end{figure}

\begin{figure}[t]
    \centering
    \includegraphics[width=\linewidth]{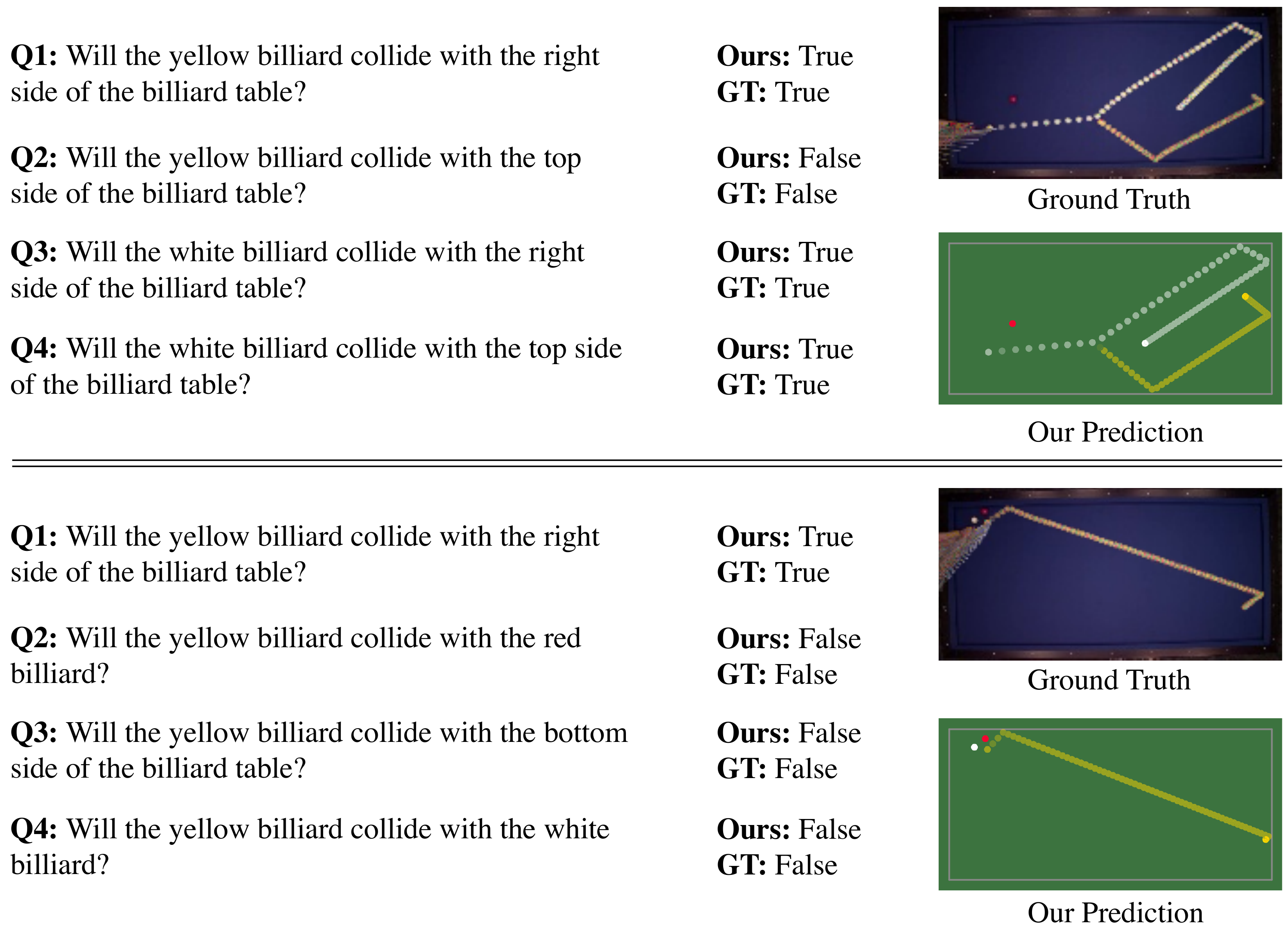}
    \caption{Visualization examples of the videos and question-answering results of our \alias~on Real-Billiards. Note that the billiard table is a chaotic system, and highly accurate long-term prediction is intractable.}
    \label{fig:examples_billiards}
\end{figure}

We show visualization examples (including failure cases) on CLEVRER~\cite{yi2019clevrer} in Fig.~\ref{fig:examples_0} and Fig.~\ref{fig:examples_1}. We also show examples on Real-Billiards~\cite{qi2020learning} in Fig.~\ref{fig:examples_billiards}. These figures show that our model can accurately learn physical parameters from video and language and perform causal simulations, predictive simulations, and counterfactual simulations for dynamic visual reasoning.
Note that the billiard table is a chaotic system, and highly accurate long-term prediction is intractable. For more failure analysis, please refer to our main paper.

\begin{table}[htbp!]
\centering
\resizebox{1\linewidth}{!}{
\begin{tabular}{cll}
\toprule
Type    & Operation & Signature \\ 
\midrule
\multirow{10}{1.4cm}{Input Operations} 
               & \texttt{Start}  & $()\rightarrow \textit{event}$  \\
               &  Returns the special ``start'' event & \\
               & \texttt{end}  & $()\rightarrow \textit{event}$  \\
               &  Returns the special ``end'' event & \\
               & \texttt{Objects}   & $()\rightarrow \textit{objects}$ \\    
               &  Returns all objects in the video &   \\
               & \texttt{Events}   & $()\rightarrow \textit{events}$\\
               & Returns all events happening in the video & \\
               & \texttt{UnseenEvents}  & $()\rightarrow \textit{events}$ \\
               & Returns all future events happening in the video & \\
\midrule
\multirow{10}{1.4cm}{Output Operations}
& \texttt{Query\_color} & $ (\textit{object}) \rightarrow \textit{color}$  \\
& Returns the color of the input object & \\ 
& \texttt{Query\_material} & $ (\textit{object}) \rightarrow \textit{material} $  \\
& Returns the material of the input objects & \\ 
& \texttt{Query\_shape} & $ (\textit{object}) \rightarrow \textit{shape} $  \\
& Returns the shape of the input objects & \\ 
& \texttt{Count} & $ (\textit{objects}) \rightarrow \textit{int} $  \\
&  Returns the number of the input objects/ events & $ (\textit{events}) \rightarrow \textit{int} $ \\
& \texttt{Exist} & $ (\textit{objects}) \rightarrow \textit{bool} $  \\
&  Returns ``yes'' if the input objects is not empty & \\
& \texttt{Belong\_to} & $ (\textit{event}, \textit{events}) \rightarrow \textit{bool} $  \\
&  Returns ``yes'' if the input event belongs to the input event sets & \\
& \texttt{Negate} & $ (\textit{bool}) \rightarrow \textit{bool} $  \\
&  Returns the negation of the input boolean & \\
\midrule
\multirow{8}{1.4cm}{Physics Operations}
& \texttt{Counterfactual\_simulation} & $ (\textit{object}) \rightarrow \textit{events}, \textit{representations} $  \\
&  Perform simulation with the object removed & \\
& \texttt{Predictive\_simulation} & $ (\textit{objects}) \rightarrow \textit{events}, \textit{representations} $  \\
&  Perform simulation after the video ends & \\
& \texttt{Apply\_heavier} & $ (\textit{object}) \rightarrow \textit{object} $  \\
&  Assign the object five times its weight before the counterfactual simulation & \\
& \texttt{Apply\_lighter} & $ (\textit{object}) \rightarrow \textit{object} $  \\
&  Assign the object one-fifth of its weight before the counterfactual simulation & \\
\midrule
\multirow{4}{1.4cm}{Object Filter Operations} 
& \texttt{Filter\_static\_concept} & $ (\textit{objects}, \textit{concept}) \rightarrow \textit{objects} $  \\
&  Select objects from the input list with the input static concept & \\ 
& \texttt{Filter\_dynamic\_concept} & $ (\textit{objects}, \textit{concept}, \textit{frame}) \rightarrow \textit{objects} $ \\
&  Selects objects in the input frame with the dynamic concept & \\ 
& \texttt{Unique} & $ (\textit{objects}) \rightarrow \textit{object} $  \\
&  Return the only object in the input list &   \\
\midrule
\multirow{18}{1.4cm}{Event Filter Operations} & \texttt{Filter\_in} & $ (\textit{events}, \textit{objects}) \rightarrow \textit{events}$  \\
&  Select incoming events of the input objects & \\ 
& \texttt{Filter\_out} & $ (\textit{events}, \textit{objects}) \rightarrow \textit{events} $  \\
&  Select existing events of the input objects & \\
& \texttt{Filter\_collision} & $ (\textit{events}, \textit{objects}) \rightarrow \textit{events} $  \\
&  Select all collisions that involve an of the input objects & \\
& \texttt{Get\_col\_partner} & $ (\textit{event}, \textit{object}) \rightarrow \textit{object} $  \\
& Return the collision partner of the input object & \\
& \texttt{Filter\_before} & $ (\textit{events}, \textit{events}) \rightarrow \textit{events} $  \\
&  Select all events before the target event & \\
& \texttt{Filter\_after} & $ (\textit{events}, \textit{events}) \rightarrow \textit{events} $  \\
&  Select all events after the target event & \\
& \texttt{Filter\_order} & $ (\textit{events}, \textit{order}) \rightarrow \textit{event} $  \\
&  Select the event at the specific time order & \\
& \texttt{Filter\_ancestor} & $ (\textit{event}, \textit{events}) \rightarrow \textit{events} $  \\
&  Select all ancestors of the input event in the causal graph & \\
& \texttt{Get\_frame} & $ (\textit{event}) \rightarrow \textit{frame} $  \\
&  Return the frame of the input event in the video & \\
& \texttt{Unique} & $ (\textit{events}) \rightarrow \textit{event} $  \\
&  Return the only event  in the input list  \\
\bottomrule
\end{tabular}}
\caption{All neuro-symbolic operations on the CLEVRER dataset~\cite{yi2019clevrer}. Our model contains five types of operations, including input, output, physics, object filter, and event filter operations.
In this table, ``order'' denotes the chronological order of an event, \eg ``First'', ``Second'' and ``Last''; ``static concept'' denotes object-level static concepts like ``Blue'', ``Cube'' and ``Metal''; ``dynamic concept'' represents object-level dynamic concepts like ``Moving'' and ``Stationary''; and ``representations'' denotes the visual features that are calculated from object trajectories.
}
\label{tb:operation}
\end{table}

\end{appendices}



\end{document}